\documentclass{article}

\usepackage{microtype}
\usepackage{graphicx}
\usepackage{booktabs} %

\usepackage{hyperref}

\usepackage{color, soul}
\usepackage{amssymb}
\usepackage{amsmath}
\usepackage{booktabs}

\usepackage{caption}
\usepackage{subfig}

\usepackage[accepted]{icml2023}

\usepackage{amsmath}
\usepackage{amssymb}
\usepackage{mathtools}
\usepackage{amsthm}

\usepackage[capitalize,noabbrev]{cleveref}

\theoremstyle{plain}

\theoremstyle{definition}

\theoremstyle{remark}

\usepackage[textsize=tiny]{todonotes}

\icmltitlerunning{\hfill Meta-Learning Mini-Batch Risk Functionals \hfill \thepage}

\begin{document}

\twocolumn[
\icmltitle{Meta-Learning Mini-Batch Risk Functionals}

\icmlsetsymbol{equal}{*}

\begin{icmlauthorlist}
\icmlauthor{Jacob Tyo}{arl,cmu}
\icmlauthor{Zachary C. Lipton}{cmu}
\end{icmlauthorlist}

\icmlaffiliation{cmu}{Machine Learning Department, Carnegie Mellon University, Pittsburgh, USA}
\icmlaffiliation{arl}{DEVCOM Army Research Laboratory, Maryland, USA}

\icmlcorrespondingauthor{Jacob Tyo}{jacob.p.tyo.civ@army.mil}

\icmlkeywords{Risk Function, Meta-Learning, deep learning}

\vskip 0.3in
]

\printAffiliationsAndNotice{\icmlEqualContribution} %
\begin{abstract}
Supervised learning typically optimizes 
the expected value risk functional
of the loss, but in many cases, 
we want to optimize for other risk functionals.
In full-batch gradient descent, 
this is done by taking gradients of a risk functional of interest,
such as the Conditional Value at Risk (CVaR) 
which ignores some quantile of extreme losses. 
However, deep learning must almost always use mini-batch gradient descent, 
and lack of unbiased estimators of various risk functionals
make the right optimization procedure unclear. 
In this work, 
we introduce a meta-learning-based method of learning 
an interpretable mini-batch risk functional
during model training, in a single shot.
When optimizing for various risk functionals, 
the learned mini-batch risk functions lead to risk reduction of up to 10\% 
over hand-engineered mini-batch risk functionals. 
Then in a setting where the right risk functional
is unknown a priori, 
our method improves over baseline by 
14\% relative ($\sim$9\% absolute).
We analyze the learned mini-batch risk functionals at different points through training, 
and find that they learn a curriculum (including warm-up periods), 
and that their final form can be surprisingly different from the 
underlying risk functional that they optimize for. 
\end{abstract}

\newcommand{\E}{\ensuremath{\mathbb{E}}}
\newcommand{\R}{\ensuremath{\mathbb{R}}}
\newcommand{\B}{\ensuremath{\mathbb{R}^b}}
\newcommand{\Ptrain}{\ensuremath{\mathbb{P}_\text{train}}}
\newcommand{\Pval}{\ensuremath{\mathbb{P}_\text{val}}}
\newcommand{\Ptest}{\ensuremath{\mathbb{P}_\text{test}}}
\newcommand{\Dtrain}{\ensuremath{\mathbb{D}_\text{train}}}
\newcommand{\Dval}{\ensuremath{\mathbb{D}_\text{val}}}
\newcommand{\Dtest}{\ensuremath{\mathbb{D}_\text{test}}}
\newcommand{\loss}{\ensuremath{\ell_{f_\theta}(X,Y)}}
\newcommand{\lossp}{\ensuremath{\ell_{f_{\theta'}}(X,Y)}}
\newcommand{\VaR}{\ensuremath{\text{VaR}_\alpha}}
\newcommand{\Xtrain}{\ensuremath{X_\text{train}}}
\newcommand{\Ytrain}{\ensuremath{Y_\text{train}}}
\newcommand{\Xval}{\ensuremath{X_\text{val}}}
\newcommand{\Yval}{\ensuremath{Y_\text{val}}}
\newcommand{\Xtest}{\ensuremath{X_\text{test}}}
\newcommand{\Ytest}{\ensuremath{Y_\text{test}}}

\section{Introduction}
\label{introduction}

The standard supervised learning procedure 
for training neural network models
is to perform mini-batch gradient descent, 
where a model is presented with a small subset of data (mini-batch)
and the model is updated with gradients derived from its outputs on that mini-batch. 
Typically, we want to minimize the expected value of the loss, 
in which case, using the expected value of the mini-batch loss 
is a clear choice. 
By the linearity of expectation, 
the gradient of the mini-batch is an unbiased estimator of 
the gradient of the dataset. 
However, 
we are often in responsible decision-making settings such as 
healthcare~\citep{patel2019human, rajpurkar2020chexaid, tschandl2020human},
credit lending~\citep{bussmann2021explainable, kruppa2013consumer}, 
and employment~\citep{raghavan2020mitigating, hoffman2018discretion}, 
where risk factors make optimizing for the expected value of the loss undesirable.
Instead, 
we often want to optimize for risk functionals such as the 
Inverted Conditional Value at Risk (ICVaR),
which ignores a percentage of the highest loss samples, 
or the trimmed risk, which ignores extreme high and low loss samples. 
In such cases, 
a mini-batch risk functional that produces unbiased gradients of 
the dataset risk functional is not available. 

\begin{figure}
    \centering
    \includegraphics[width=0.5\textwidth]{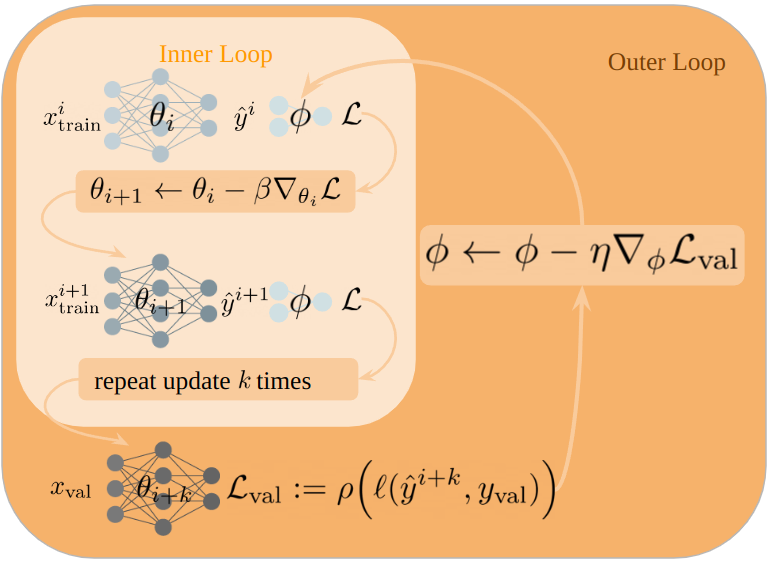}
    \caption{A high-level description of our meta-learning procedure. Given a model parameterized by $\theta$, this procedure learns a mini-batch risk functional, parameterized by $\phi$, that is a convex combination of the mini-batch losses. Trained in a single shot alongside the model, the learned mini-batch risk functional provides a means of better optimizing for dataset risk functionals of interest ($\rho$) and can be used to automatically detect and deal with some forms of distribution shift.}
    \label{fig:learningRiskFn}
\end{figure}

This is the first setting we focus on:
determining the right mini-batch risk functionals 
to use for the optimization of various dataset risk functionals.
But we also study a second, more challenging setting, 
where given a problem where a risk functional 
could be useful but is \emph{unknown} a priori, 
can we meta-learn the right mini-batch risk 
functional for maximum performance?

In this paper, 
we introduce a meta-learning-based method of learning 
an interpretable mini-batch risk functional
during model training, in a single shot,
that is effective in both aforementioned settings. 
When optimizing for various risk functionals, 
the learned mini-batch risk functions lead to risk reduction of up to 10\% 
over hand-engineered mini-batch risk functions. 
Then in the presence of label noise, 
where risk functionals can be useful but the right one is unknown, 
our method improves over baseline by 
over 14\% relative ($\sim$9\% absolute) among 50\% random labels, 
even when given \emph{no} noise-free data.
Our learned mini-batch risk functions 
are restricted to taking a convex
combination of mini-batch loss quantiles, 
and are therefore clearly interpretable. 
We analyze the learned mini-batch risk functionals at different points through training, 
and find that they learn curriculums (including warm-up periods), 
and that their final form can be surprisingly different from the 
underlying risk functional that they are optimizing for.

\begin{table*}[hbt]
\small
\centering
\caption{Definitions and interpretations of common risk functionals. $F_f(\loss)$ represents the CDF of $\loss$ and $\text{VaR}_\alpha = 100 \times \alpha$-percentile. For more discussion on these risk functionals, see \citet{wongriskyzoo}.}
\label{tab:riskfunctions}
\begin{tabular}{@{}ccc@{}}
\toprule
Risk Functional      & Expression & Interpretation                                                                                  \\ \midrule
Expected Value     & $\E[\loss]$ & expected loss \\ \addlinespace[0.5em]
CVaR               & $\E[\loss | \loss \geq \VaR (\loss)]$ & expectation of losses exceeding the $100\cdot \alpha$ percentile \\ \addlinespace[0.5em]
Inverted CVaR      & $\E[\loss | \loss \leq \VaR(\loss)]$           & expectation of losses below the $100 \cdot \alpha$ percentile                              \\ \addlinespace[0.5em]
Human-aligned & $\E[\loss w(F_f(\loss))]$ & weighting function that overweights extreme losses \\ \addlinespace[0.5em]
Mean-Variance      & $\E[\loss] + c \cdot \text{Variance}[\loss]$           & expected loss penalized by its variance                                                         \\ \addlinespace[0.5em]
Trimmed Risk       & $\E[\loss|F_f(\loss) \in [\alpha, 1-\alpha]]$ & ignore extreme losses                                                                           \\ \bottomrule
\end{tabular}
\end{table*}

Our core contributions are: 
\begin{itemize} \itemsep0em 
    \item meta-learning-based method for learning an interpretable mini-batch risk functional, in a single shot
    \item empirical study revealing that our learned mini-batch risk functionals better optimize for risk functionals of interest, reducing risk by up to 10\%
    \item experimental demonstration that our learned mini-batch risk functionals can learn effective functionals even when a useful hand-crafted risk function is unknown a priori, improving over baseline by 14\% relative ($\sim$9\% absolute) among 50\% random labels
    \item analysis of the learned mini-batch risk functionals revealing their surprising difference from hand-crafted risk functionals, and interesting dynamics such as automatic curriculum development.
\end{itemize}

\begin{table*}[hbt]
\small
\centering
\parbox{0.92\textwidth}{
\caption{Definitions and interpretation of common distribution shift settings. $\Ptrain(X,Y)$ represents the joint distribution over $X$ and $Y$ of the training distribution.} \label{tab:distShifts}}
\begin{tabular}{@{}ccc@{}}
\toprule
Distribution Shift         & Expression                                                                                                & Interpretation                                         \\ \midrule
General Distribution Shift & $\Ptrain(X,Y) \not= \Ptest(X,Y)$                                                                          & The training and testing distributions are not the same \\ \addlinespace[0.5em]
Covariate Shift            & \begin{tabular}[c]{@{}c@{}}$\Ptrain(X) \not= \Ptest(X)$, but \\ $\Ptrain(Y|X) = \Ptest(Y|X)$\end{tabular} & Class balance changes from train to test set           \\ \addlinespace[0.5em]
Label Shift                & \begin{tabular}[c]{@{}c@{}}$\Ptrain(Y) \not= \Ptest(Y)$, but\\ $\Ptrain(X|Y) = \Ptest(X|Y)$\end{tabular}  & A common example of this shift is noisy labels       \\ \bottomrule
\end{tabular}
\end{table*}

\section{Risk Sensitive Learning}
\label{risksensitivelearning}

First, we setup the problem of risk-sensitive learning.
Specifically, 
given training data $\{ X_i, Y_i \}_{i=1}^n$ 
comprising the dataset \Dtrain \ 
(which is drawn from the distribution \Ptrain)
where $C$ is the number of classes, 
$b$ is the batch size, 
$d$ is the dimensionality of the input,
$X \in \mathcal{X} \subseteq \mathbb{R}^d$ and $Y \in \mathcal{Y} \subseteq \mathbb{R}^C$,
a loss function $\ell : \mathcal{Y} \times \mathcal{Y} \to \mathbb{R}^{C}$, 
a risk functional $\rho : \mathbb{R}^b \to \mathbb{R}$ 
maps a loss distribution to a real value, 
and a hypothesis class $\Theta$, 
risk-sensitive learning optimizes for
\begin{align}
     \theta^* \in \arg \underset{\theta \in \Theta}{\min} \ \rho (\loss),  %
\end{align}
where $\loss := \ell(f_{\theta}(X), Y)$ is the loss 
under model $f$ parameterized by $\theta$.  
Furthermore, 
we will use \Xtrain, \Ytrain \ and \Xval, \Yval \ 
to represent mini-batches of training and validation data respectively. 

This formulation encapsulates traditional learning, 
in which case the risk functional is set to be the expected value: $\rho = \mathbb{E}$. 
Risk-sensitive learning explores the use of different risk functionals $\rho$, 
whether for training models when all samples 
should \emph{not} be treated equally
(i.e. fairness, dangerous actions, etc.),
or to address specific types of distribution shifts. 
Table~\ref{tab:riskfunctions} details common risk functionals 
along with their formal definition and interpretation. 
Table~\ref{tab:distShifts} enumerates common
types of distribution shifts. 
\section{Learning Risk Functionals}
\label{sec:learningRiskFns}

As a final note before introducing our method in detail, 
our method requires two datasets during training. 
We use the terms training, validation, hyper-validation, and test
to refer to each of the dataset splits. 
Specifically, given a train and a test set, 
we split the train set into a train, a validation, 
and a hyper-validation set randomly as 90\%/5\%/5\% of the original train set,
leaving the test set unaltered. 
From this point on, 
when the validation data is mentioned, 
we are referring to that specific split of the training data.
For hyperparameter selection, we leverage the hyper-validation set. 

In this work, 
we propose a method to automatically learn mini-batch risk functionals, 
directly from data,
based on a MAML-like~\citep{finn2017model} meta-learning formulation. 
Specifically, the mini-batch risk functional $g_\phi$ is
implemented as a linear layer, restricted to taking a convex combination
of mini-batch loss quantiles.
We make no assumption on the form of the model $f$, 
other than to assume that it is parameterized by some parameter vector $\theta$, 
and that gradient-based learning techniques can be used. 
To optimize $g_\phi$, 
we first set $\theta' = \theta$
and then repeat the following update for some number of \emph{inner steps}
\begin{align}
\label{eq:innerstep}
    \theta' \gets \theta' - \beta \nabla_{\theta'} g_\phi \Big( \ell_{f_{\theta'}}(\Xtrain, \Ytrain ) \Big),
\end{align}
where $\beta$ is the inner learning rate, resulting in an updated $f_{\theta'}$. 
We then take an \emph{outer step} by passing a mini-batch of validation 
data through $f_{\theta'}$, 
and then differentiating through 
this entire process with respect to $\phi$. 
Therefore, the update for $\phi$ is: 
\begin{align}
\label{eq:outerstep}
    \phi \gets \phi - \eta \nabla_\phi \rho \Big( \ell_{f_{\theta'}}(\Xval, \Yval ) \Big).
\end{align}
Notice that this update only depends on $\phi$ via the derivation of $\theta'$,
which is critical in preventing degenerate gradients. 
If gradients of $\phi$ are calculated in the inner steps, 
or if calculated after a single update, 
then $\phi$ can push the loss arbitrarily low, 
regardless of model output. 
Finally, to complete this outer step and continue training, 
the model parameters are reset to $\theta$, 
a new mini-batch of training data is drawn, 
and the model is updated with 
\begin{align}
    \theta \gets \theta - \beta \nabla_{\theta} g_\phi \Big( \ell_{f_{\theta}}(\Xtrain, \Ytrain ) \Big). 
\end{align}
In summary, we minimize the following objective at each time-step
\begin{align}
    g_{\phi^*} &\in \arg \underset{g \in \mathcal{G}}{\min} \ \rho \Bigg(\ell \Big(f_{\theta'}(\Xval), \Yval \Big) \Bigg)  \\ 
    \text{where} \quad f_{\theta'} &\in \arg \underset{f \in \mathcal{F}}{\min} \ \ell \Bigg( g_\phi \Big( f_\theta(\Xtrain), \Ytrain \Big) \Bigg).
\end{align}

Algorithm~\ref{alg:learnedRiskFn} details this optimization procedure in 
pseudocode. 
To point out the critical details, 
note that in line 6, the loss for each data point in the mini-batch 
is sorted from largest to smallest. 
Sorting from largest to smallest is not particularly important, 
but it is critical that the batch losses are sorted in some fashion before 
passing them through $g$. 
This allows for learning meaningful relations 
with respect to the loss magnitudes instead of the random mini-batch order.
In line 8, the gradients for the learnable risk functional are dropped. 
This is an important implementation note, 
as without dropping these gradients they 
will be accumulated as part of the outer update, 
resulting in undesirable behavior 
(i.e. the gradients in Equation~\ref{eq:innerstep} will also be calculated 
with respect to $\phi$ as well). 
Then in line 12, we use a hand-engineered risk functional $\rho$
to calculate the validation loss. 
This optimization pipeline can be interpreted as
learning a loss reduction functional that minimizes a desired risk on the validation set, 
given only information from the current mini-batch of training data.

Learning a risk functional can have 
an effect similar to changing learning rates. 
For example, if we just use $2*\E[\loss]$ as the mini-batch risk functional, 
then this is equivalent to using the traditional expected value
but also doubling the learning rate. 
Therefore, we restrict our function class to be a convex combination 
of the loss for each data point to eliminate this effect. 
This comes with the additional benefit of being able to 
represent all of the aforementioned risk functionals (up to scaling factors)
and provides straightforward interpretation. 
This is enforced by using the softmax of the weights 
of the learned risk functional linear layer. 
After training, 
the learned model is evaluated in exactly the same manner as any other model.

\begin{algorithm}[htb]
\caption{Pseudocode for learning risk functionals.}
\label{alg:learnedRiskFn}
\begin{algorithmic}[1]
\REQUIRE initial model parameters $\theta$ and $\phi$, inner learning rate $\beta$, outer learning rate $\eta$, validation risk functional $\rho$, loss function without reduction $\ell$, a training dataset $\Ptrain$, and a validation dataset $\Pval$
\WHILE{not done}
    \STATE $f_{\theta'} = f_\theta$
    \FOR{inner steps}
        \STATE $\Xtrain, \Ytrain \sim \Dtrain$ \ \#sample batch of train data
        \STATE $\hat y = f_{\theta'}(\Xtrain)$ \ \#Get model output
        \STATE $l_\text{sorted} = \text{sort}( \ell(\hat y, \Ytrain ) )$ \ \#Compute and sort Loss
        \STATE $l = g_\phi(l_\text{sorted})$ \ \#get risk 
        \STATE $\theta' = \theta' - \beta \nabla_{\theta'} l $ \ \#adapt $\theta'$
    \ENDFOR
    \STATE $\Xval, \Yval \sim \Dval$ \ \#sample validation data
    \STATE $\hat y = f_{\theta'}(\Xval)$ \ \#get updated model output
    \STATE $l = \rho(\ell(\hat y, \Yval)$ \ \#Compute validation loss
    \STATE $\phi = \phi - \eta \nabla_{\phi} l$ \ \#Update $g$
    \STATE $\Xtrain, \Ytrain \sim \Dtrain$ \ \#sample new train data
    \STATE $\hat y = f_\theta(\Xtrain)$ \ \#get model output using original $\theta$
    \STATE $\theta = \theta - \beta \nabla_\theta g_\phi(\ell(\hat y, \Ytrain)) $ \ \#update original $\theta$
\ENDWHILE
\end{algorithmic}
\end{algorithm}

\begin{table*}[htb]
\centering
\caption{Given a risk functional to minimize (column 1), this table compares the risk achieved when optimizing a model using expected value (column 2), a model optimized with the corresponding risk functional applied per batch (column 3), a warm-started model fine-tuned with the corresponding risk functional applied per mini-batch (column 4), and a model that learns a mini-batch risk functional (column 5). The lowest-risk entries are bolded. The reported metrics are averaged over 5 runs, with the standard deviation reported in parentheses.}
\label{tab:learned_riskfns}
\begin{tabular}{@{}ccccc@{}}
\toprule
    $\rho$     & Expected Value                & batch $\rho$       & batch $\rho$ with warm-up & Learned      \\ \midrule
Expected Value     &  \textbf{0.2679 (0.00466)}    &  \textbf{0.2679 (0.00466)}   &   \textbf{0.2679 (0.00466)}      & \textbf{0.2776 (0.00473)}   \\
CVaR     & 2.301 (0.0317)     & 1.917 (0.00962)    & 1.773 (0.00976)        & \textbf{1.721 (0.0156)}   \\
ICVaR    & 1.459e-6 (1.42E-7) & 8.372E-5 (3.26E-5) & 1.430e-5 (1.59e-6)     & \textbf{1.341e-7 (3.23e-8)}           \\
Human    & 0.5953 (0.00610)   & \textbf{0.576 (0.00762)}   & \textbf{0.5753 (0.00867)}        & 0.6055 (0.0169)   \\
Mean Var & 0.3420 (0.00867)    & \textbf{0.3288 (0.0109)}    & \textbf{0.3270 (0.00332)}        & 0.3477 (0.00396) \\
Trimmed  & 0.04801 (0.00132)  & \textbf{0.04401 (0.00153)}  & 0.05036 (0.00334)      & \textbf{0.04602 (0.0026)}  \\ \bottomrule
\end{tabular}
\end{table*}

\begin{table*}[htb]
\centering
\caption{Given a risk functional to minimize (column 1), this table compares the risk of a model optimized with the corresponding risk functional applied per mini-batch (column 2), a warm-started model fine-tuned with the corresponding risk functional applied per mini-batch (column 3), and a model that learns a mini-batch risk function (column 4). The bolded entries represent the highest accuracy models. The reported metrics are averaged over 5 runs, with the standard deviation reported in parentheses.}
\label{tab:learned_riskfns_acc}
\begin{tabular}{@{}cccc@{}}
\toprule
     $\rho$    & batch $\rho$              & batch $\rho$ w/warm-up  & Learned             \\ \midrule
Expected Value     &     \textbf{91.12 (0.290)}      &   -      & \textbf{90.76 (0.470)} \\
CVaR     & 68.208 (1.02)           & 79.14 (0.447)          & \textbf{85.46 (0.347)} \\
ICVaR    & 16.78 (1.48)          & 89.26 (0.215)          & \textbf{91.07 (0.0764)} \\
Human    & \textbf{90.51 (0.141)}  & \textbf{90.32 (0.416)} & \textbf{90.25 (0.530)}          \\
Mean Var & \textbf{91.17 (0.161)} & 90.80 (0.204)          & 90.75 (0.269)           \\
Trimmed  & 89.05 (0.134)          & 89.37 (0.172)          & \textbf{90.52 (0.149)} \\ \bottomrule
\end{tabular}
\end{table*}

\section{Experimental Evaluation}
\label{sec:experiments}

Unless otherwise noted, 
all experiments were performed on the CIFAR10~\citep{krizhevsky2009learning} dataset
with a Resnet-18~\citep{he2016deep} model. 
In every setting,
we optimize the model for 20,000 steps, 
with a one-cycle learning rate schedule~\citep{smith2019super} 
that starts with a learning rate of 0.005, 
increasing to a maximum of 0.1,
and then annealing back down to 5e-6. 
Random cropping and flipping are used for data augmentation, 
and the models are trained using SGD with a momentum of 0.9, 
weight decay of 5e-4, and a batch size of 100. 
These hyperparameters were selected as they were the best performing 
from a grid search 
on the hyper-validation set. 

The learned mini-batch risk functionals 
are trained using Algorithm~\ref{alg:learnedRiskFn}, 
using the same aforementioned setup, 
but with the optimizer 
set to be Adam with a learning rate of 0.001
5 inner steps (Line 3 in Algorithm~\ref{alg:learnedRiskFn}) are used, 
and a meta-validation batch size of 2000 (Line 10 in Algorithm~\ref{alg:learnedRiskFn}).
In all of these experiments, 
the risk functionals are learned in a single shot along with the model.
We report the performance of each model after the metric of interest
does not improve after 5 epochs, or the 20,000 step limit is reached. 

In some settings, we refer to \emph{warm-starting} a model.
By warm-starting, we mean using all of the same hyperparameters mentioned above, 
but for 10,000 steps 
(including updating the learning rate schedule for this timeframe) 
we do normal supervised learning with the expected value risk functional. 
We then fine-tune the warm-started model for another 10,000 steps 
(again with the only change being the new annealing schedule),
using the risk functional of interest. 
This procedure allows for the smoothest transfer between 
the learning objectives,
produced the best results, 
and maintains consistency in the number of optimization steps used.

\subsection{Optimizing for various risk functions}
\label{sec:optimizingRiskFns}

We first set out to understand the mini-batch risk functionals that are
uncovered when optimizing for various hand-crafted dataset risk functionals.
Given a risk functional of interest 
(such as any from Table~\ref{tab:riskfunctions}), 
we compare the following methods of optimizing for it:
1) train normally with expected value as the risk functional,
2) train with the risk functional of interest applied at each mini-batch,
3) warm-start a model 
and then fine-tune using the risk functional applied at each mini-batch, 
and 4) learn a batch-level risk functional with our method. 
For each risk functional in Table~\ref{tab:riskfunctions}, 
we report the average and standard deviation across 5 runs. 
Table~\ref{tab:learned_riskfns} detail these losses, 
and Table~\ref{tab:learned_riskfns_acc} detail the accuracy's. 

\begin{figure*}[htb]
    \centering
    \subfloat[CVaR]{\label{fig:CVAR}\includegraphics[width=0.2\textwidth]{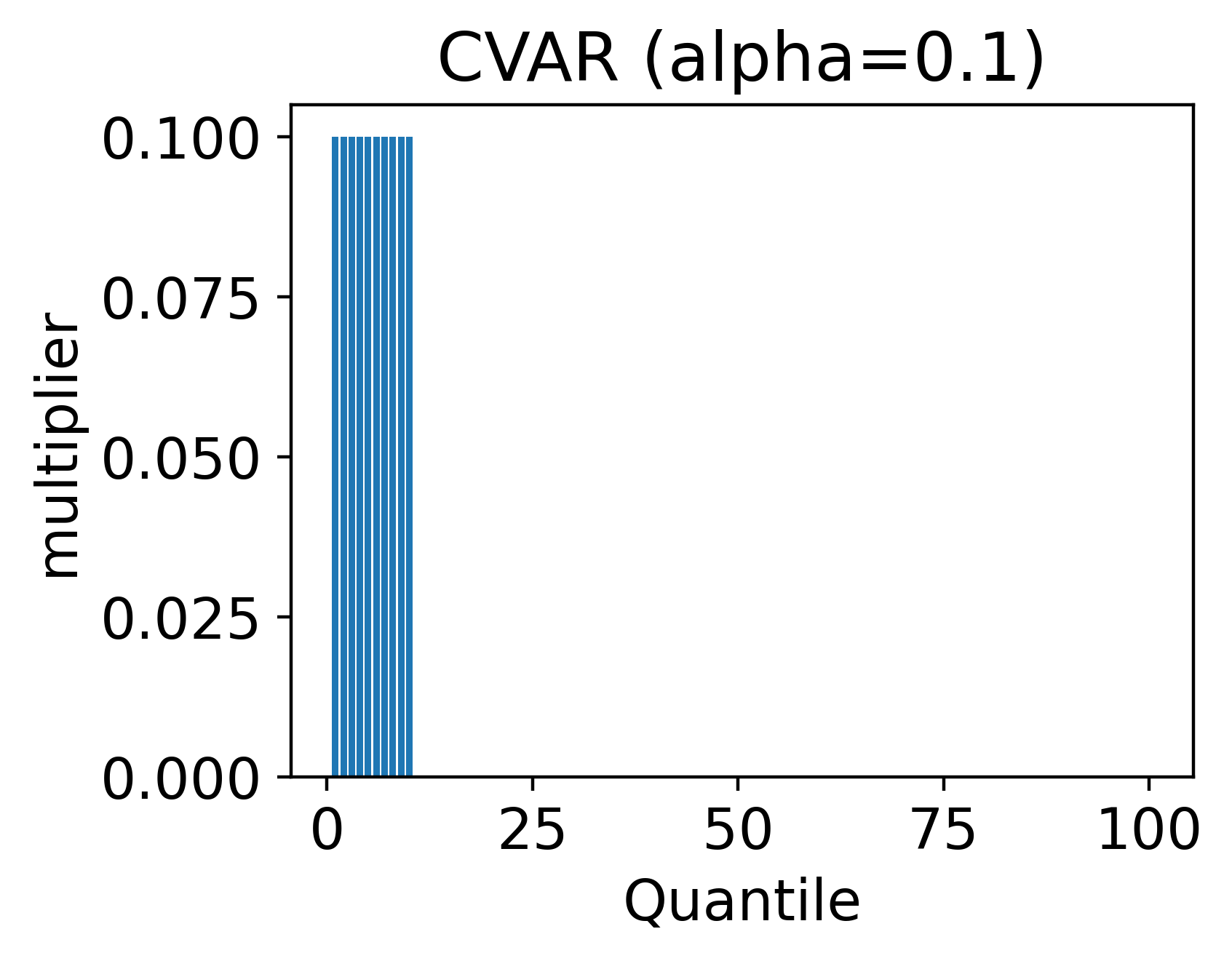}}
    \subfloat[t=100]{\label{fig:CVAR0}\includegraphics[width=0.2\textwidth]{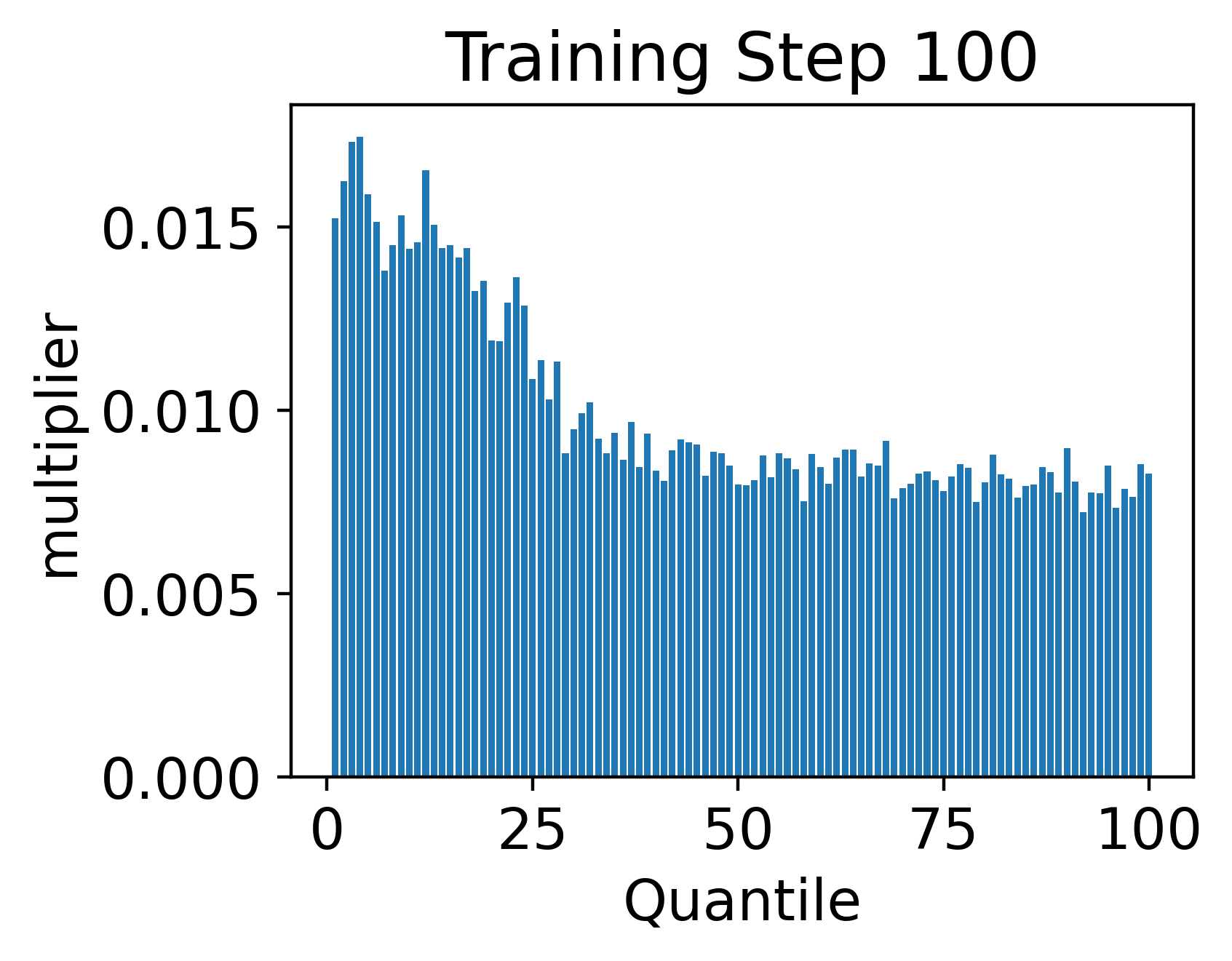}}
    \subfloat[t=1000]{\label{fig:CVAR2}\includegraphics[width=0.2\textwidth]{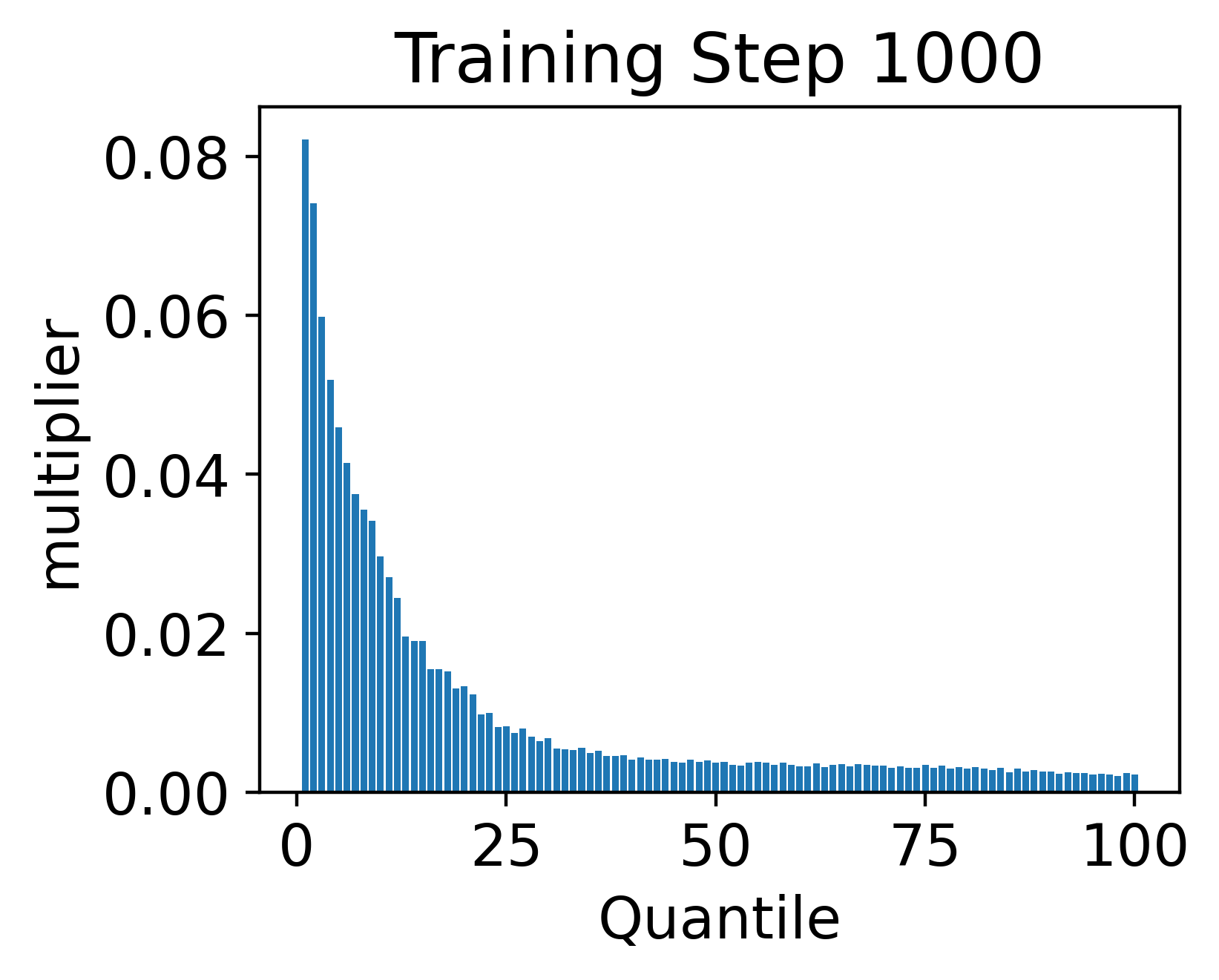}}
    \subfloat[t=10000]{\label{fig:CVAR5}\includegraphics[width=0.2\textwidth]{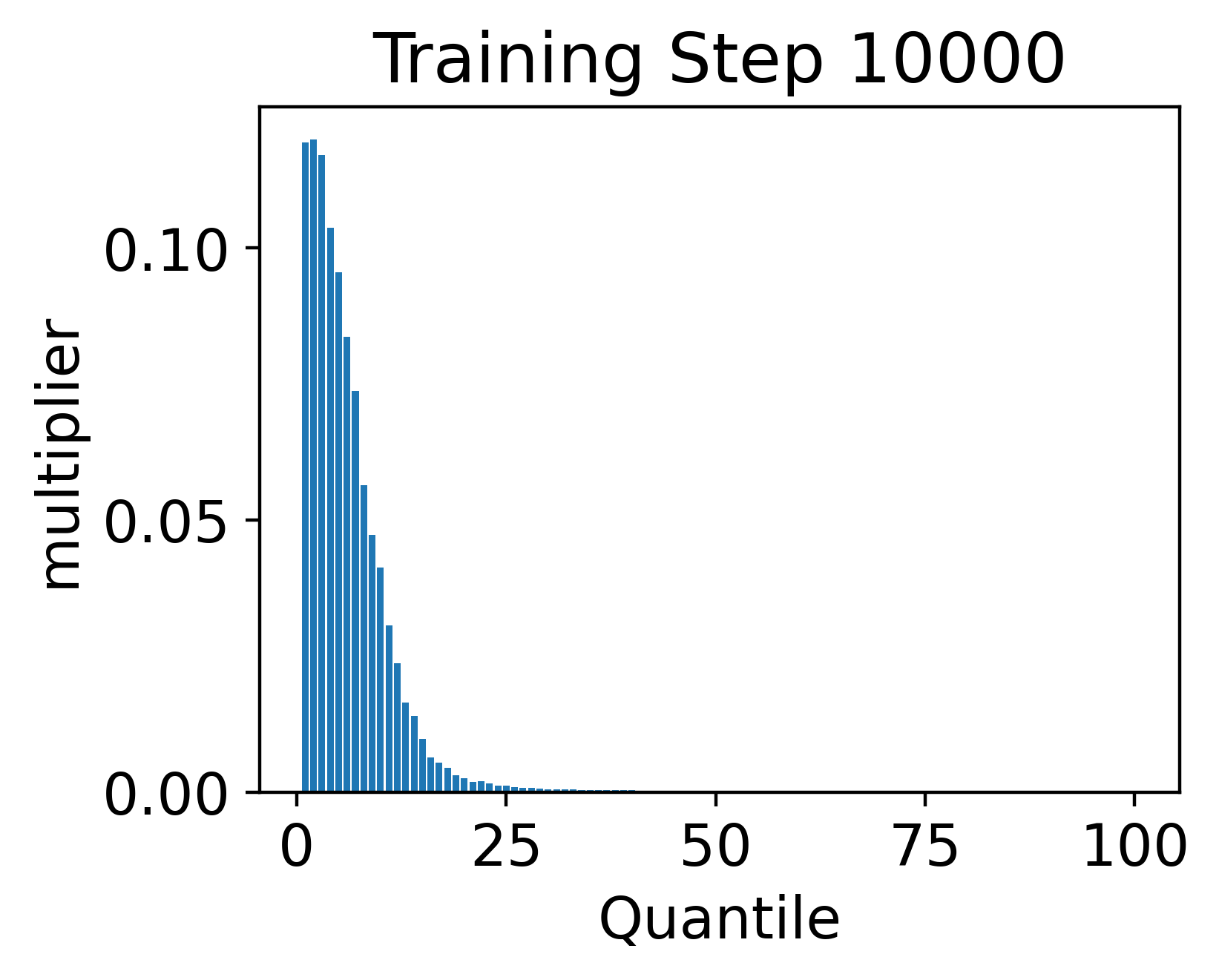}}
    \subfloat[t=20000]{\label{fig:CVAR6}\includegraphics[width=0.2\textwidth]{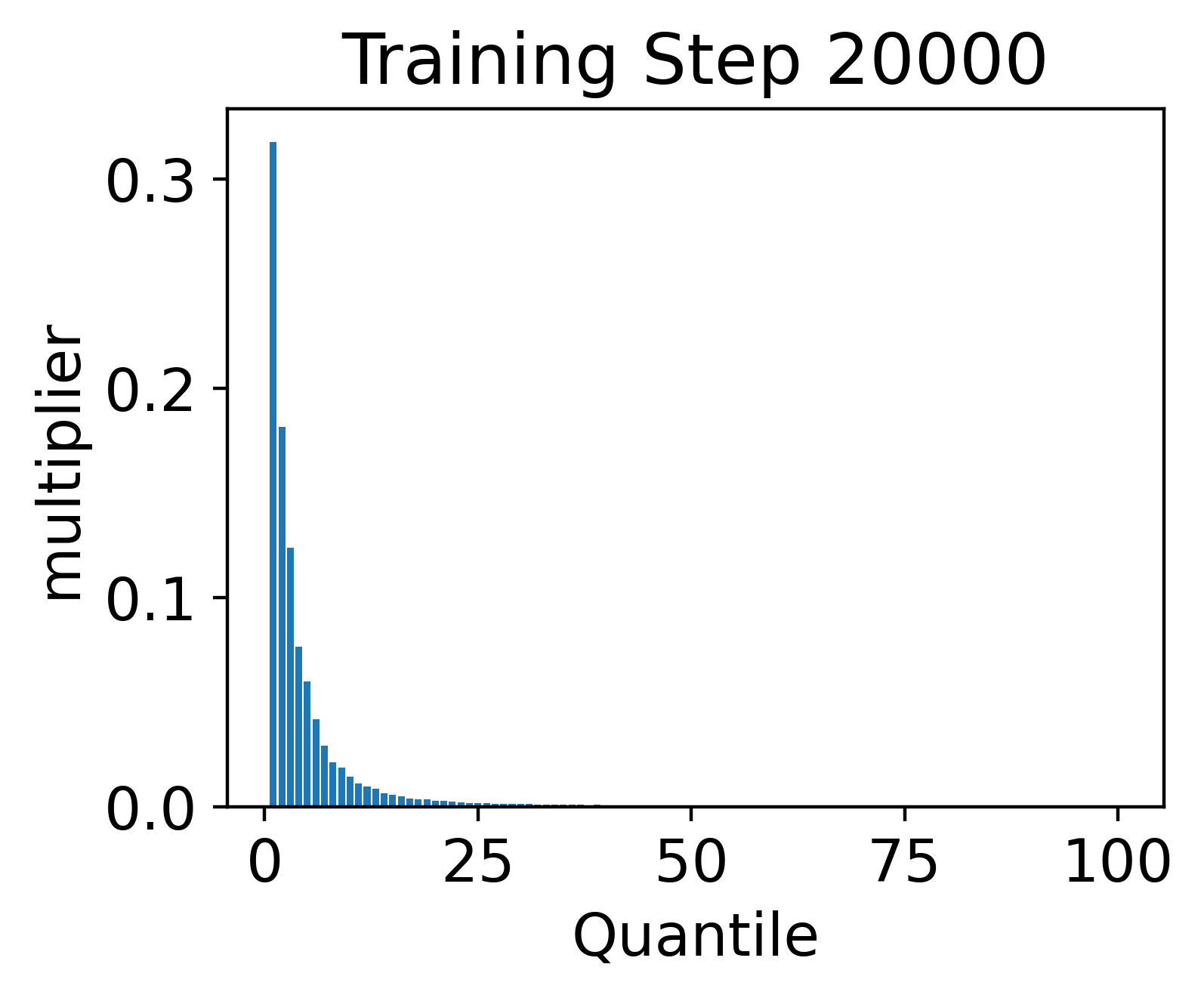}}
    \caption{Visualizations of the mini-batch risk functional learned when optimizing for the CVaR ($\alpha = 0.1$) risk of the CIFAR-10 dataset, using a Resnet-18 model. Subfigure \ref{fig:CVAR} indicates the dataset-level risk functional we are optimizing for, and the remaining figures are the learned mini-batch risk functionals at different training time steps (indicated by the subfigure caption). The y-axis indicates the multiplier applied to the corresponding quantile (sorted by magnitude) of each batch. The left side of each plot corresponds to the largest loss in the mini-batch, and the right side of each plot corresponds to the smallest loss in the mini-batch. The learned mini-batch risk functional learns a warm-up schedule before finally focusing on the highest loss samples.}
    \label{fig:CVAR_RiskFns}
\end{figure*}

\begin{figure*}[htb]
    \centering
    \subfloat[ICVaR]{\label{fig:ICVAR}\includegraphics[width=0.2\textwidth]{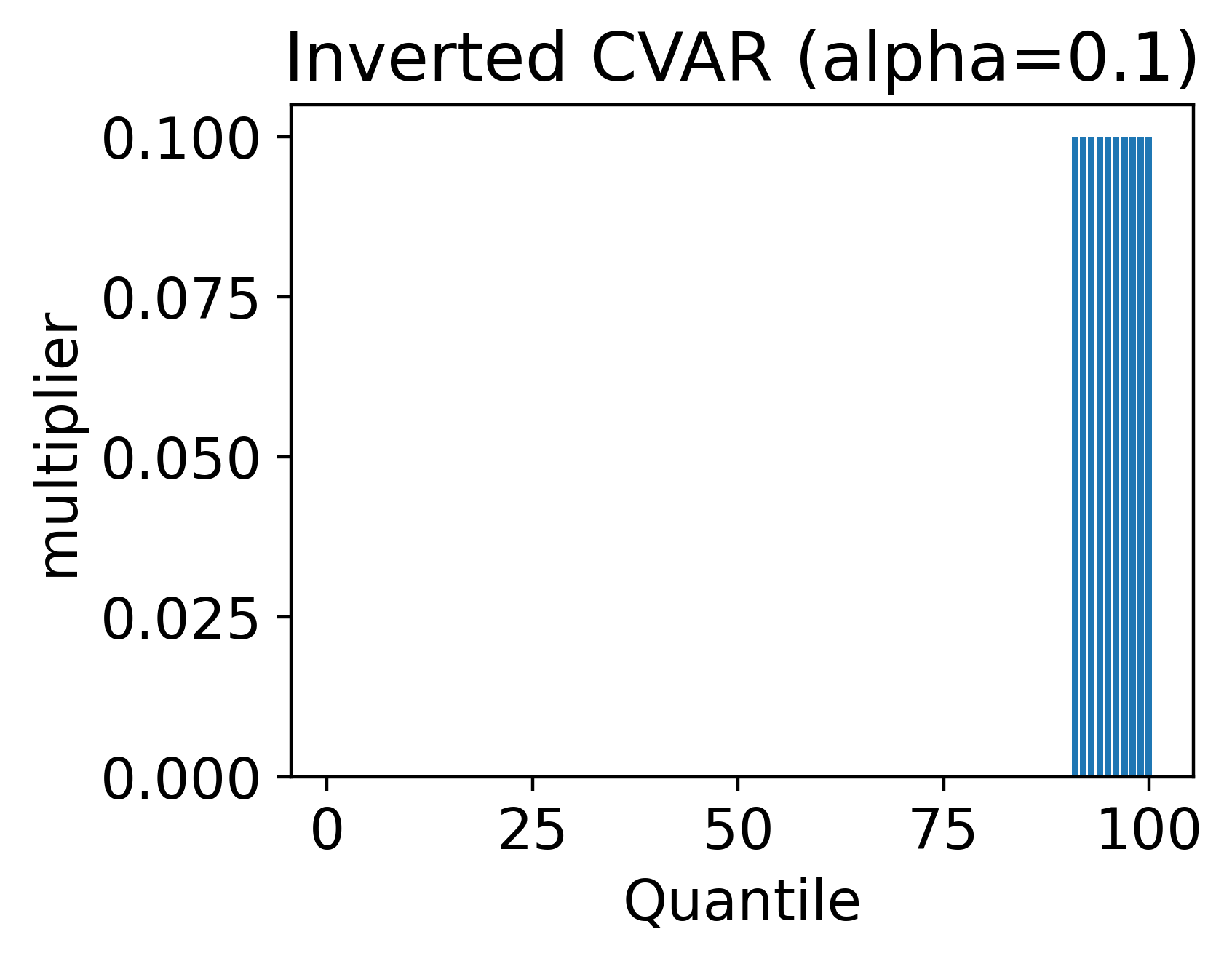}}
    \subfloat[t=100]{\label{fig:ICVAR0}\includegraphics[width=0.2\textwidth]{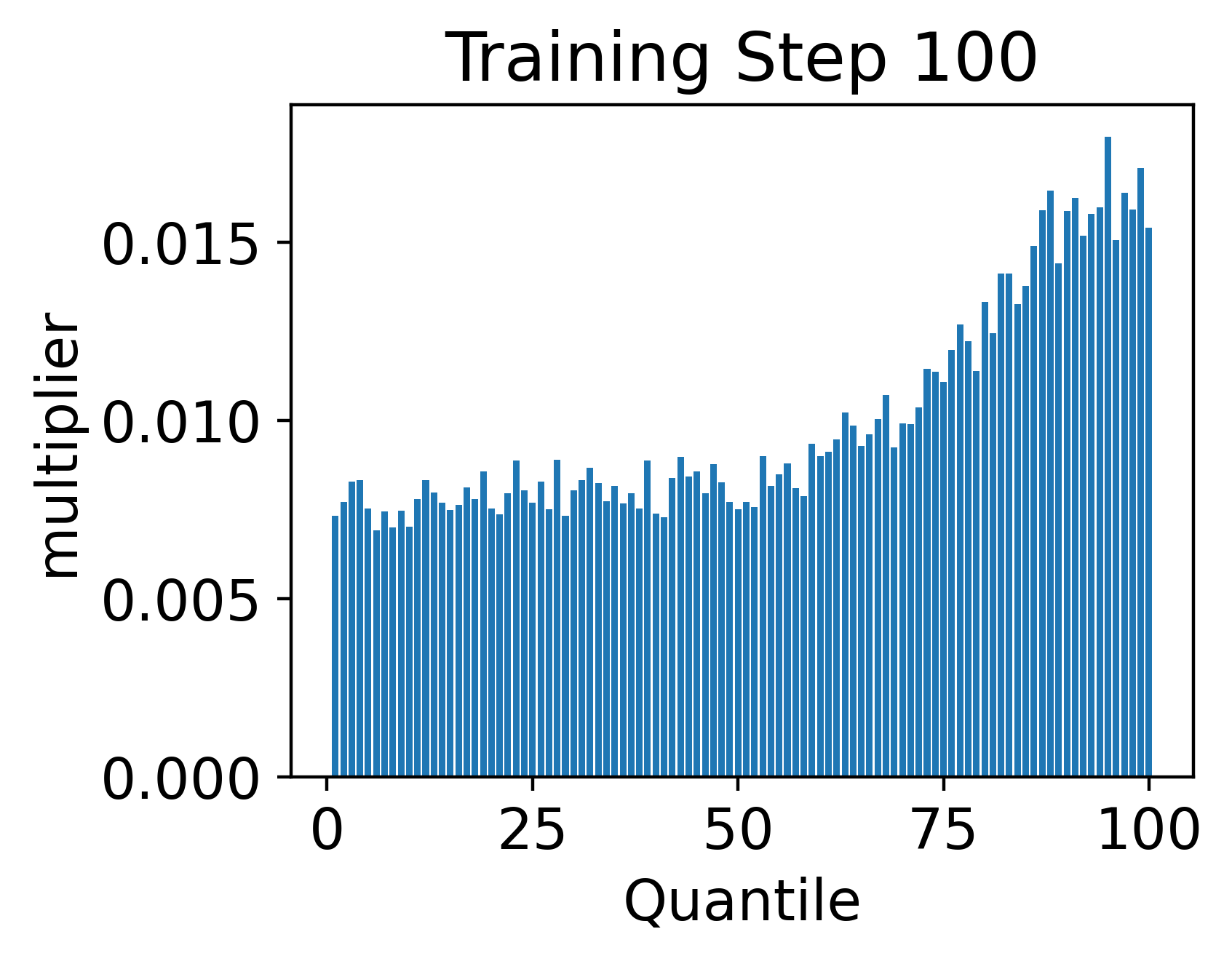}}
    \subfloat[t=1000]{\label{fig:ICVAR2}\includegraphics[width=0.2\textwidth]{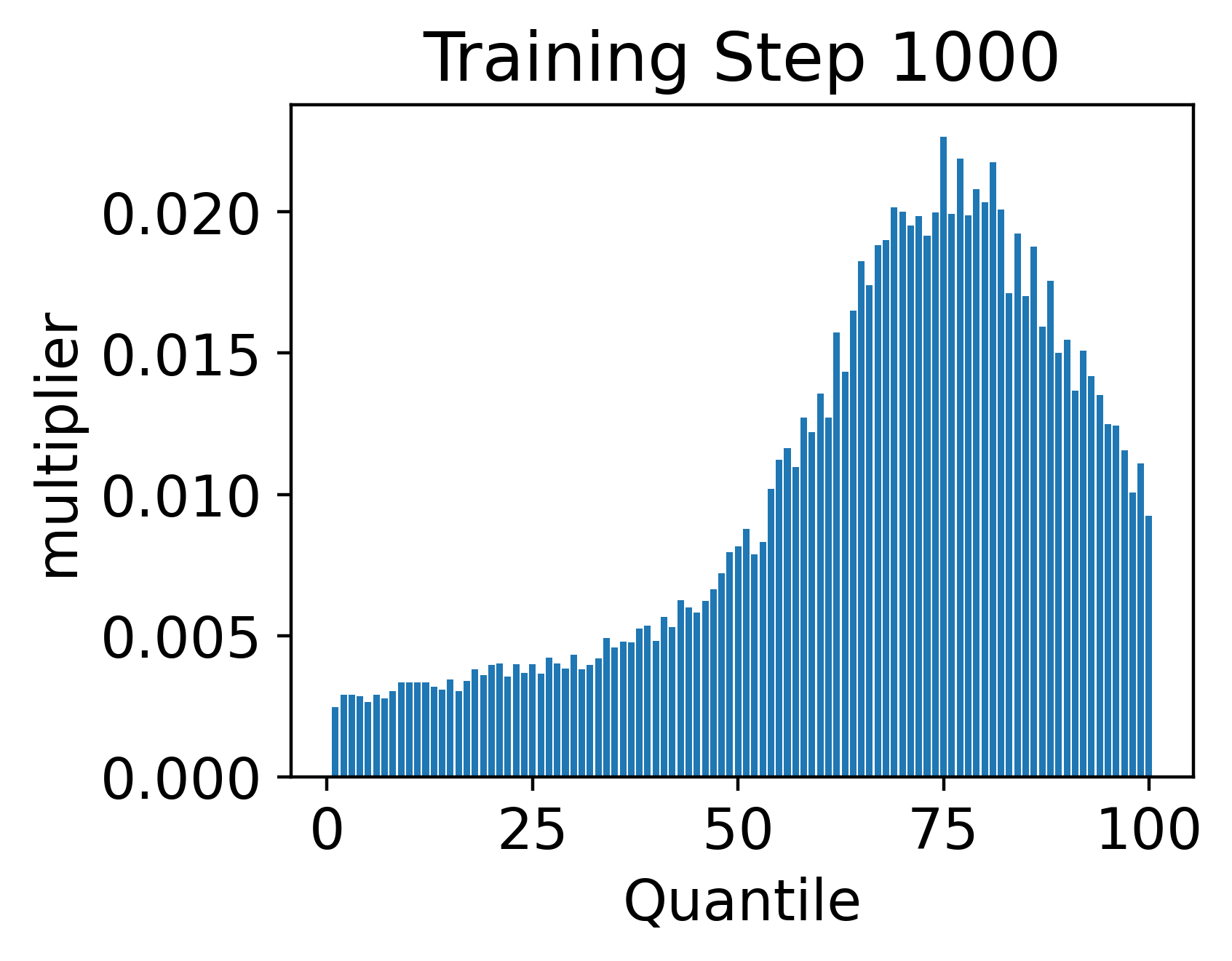}}
    \subfloat[t=10000]{\label{fig:ICVAR5}\includegraphics[width=0.2\textwidth]{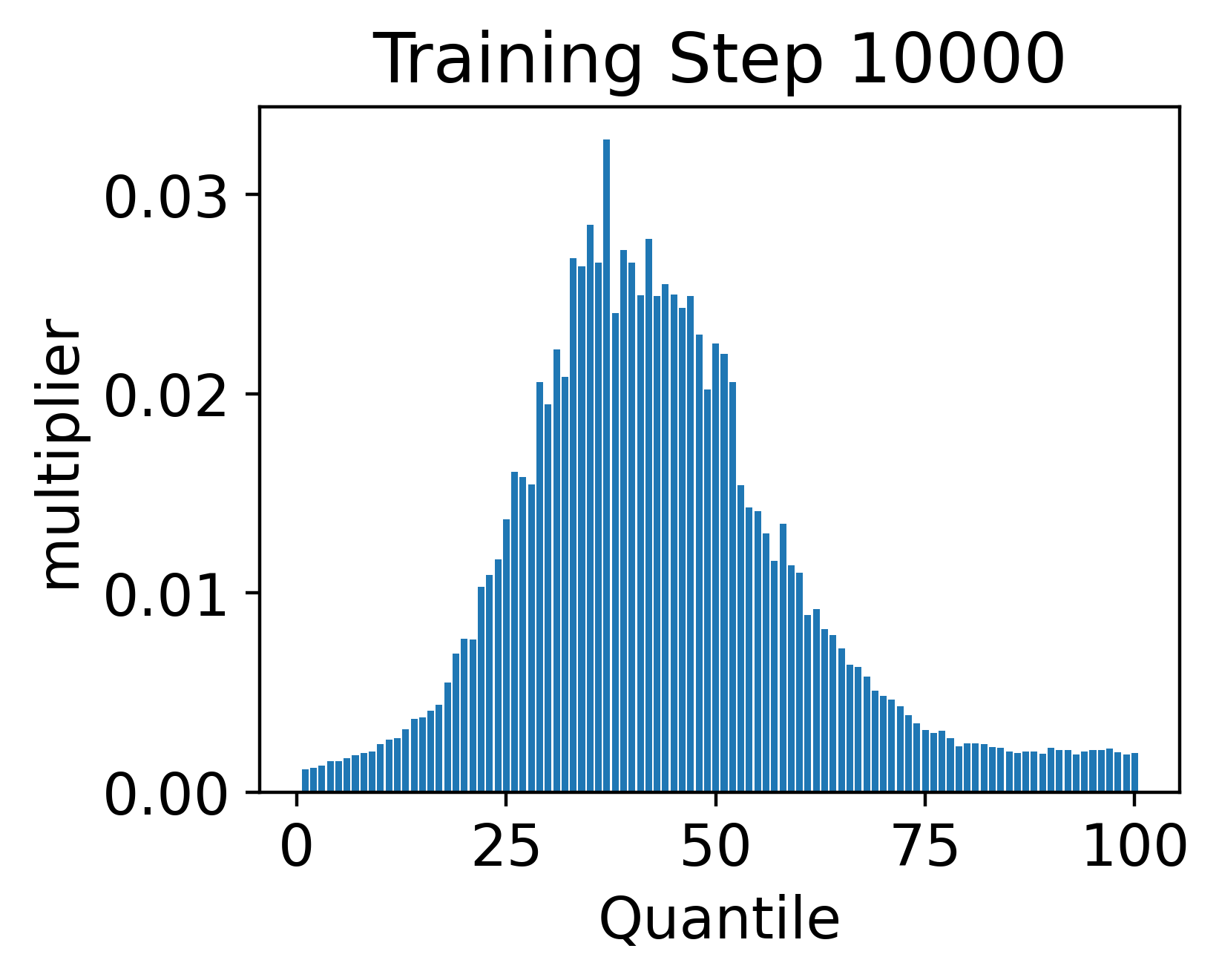}}
    \subfloat[t=20000]{\label{fig:ICVAR6}\includegraphics[width=0.2\textwidth]{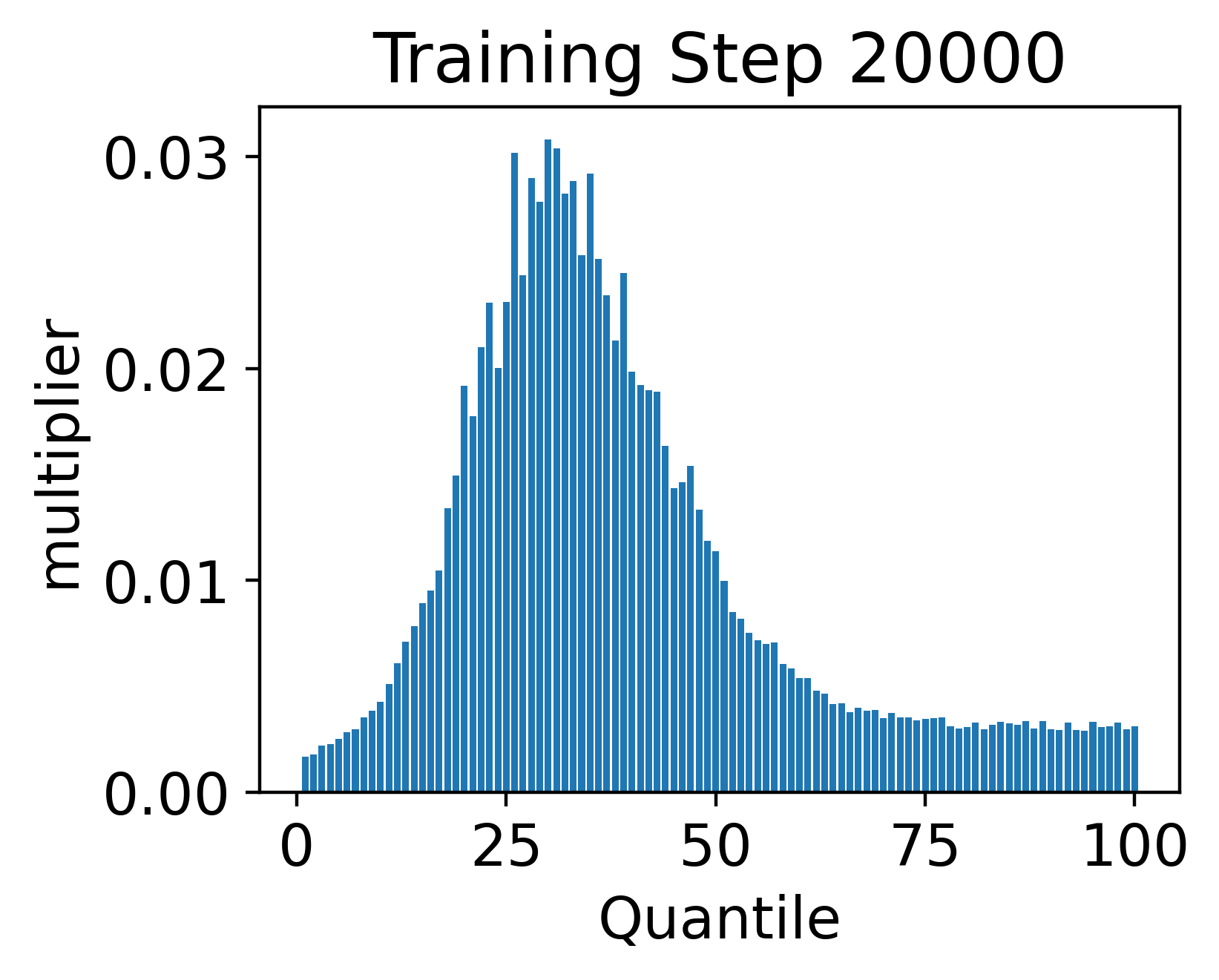}}
    \caption{Visualizations of the mini-batch risk functional learned when optimizing for the ICVaR ($\alpha = 0.1$) risk of the CIFAR-10 dataset, using a Resnet-18 model. Subfigure\ref{fig:ICVAR} indicates the dataset-level risk functional we are optimizing for, and the remaining figures are the learned risk functional at different training time steps (indicated by the subfigure caption). The y-axis indicates the multiplier applied to the corresponding quantile (sorted by magnitude) of each mini-batch. The left side of each plot corresponds to the largest loss in the mini-batch, and the right side of each plot corresponds to the smallest loss in the mini-batch. The learned risk functional learns a warm-up schedule, and although initially focusing on the lower mini-batch losses, the emphasis is placed more on middle quantiles of losses later in training, although the ICVaR loss penalizes only the lowest loss samples.}
    \label{fig:ICVAR_RiskFns}
\end{figure*}

\begin{figure*}[htb]
    \centering
    \subfloat[Trimmed Risk]{\label{fig:trimmed}\includegraphics[width=0.2\textwidth]{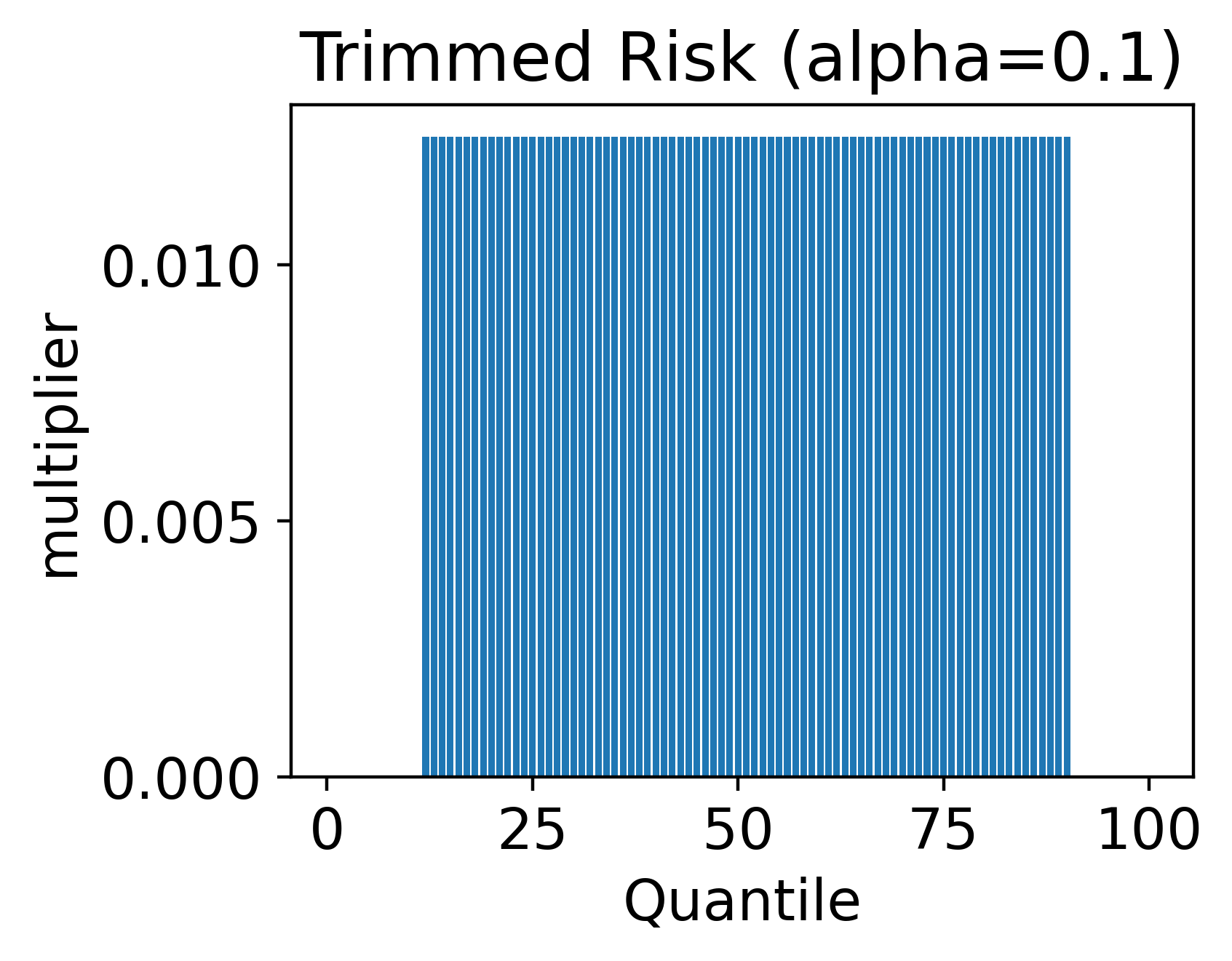}}
    \subfloat[Learned t=100]{\label{fig:trimmed0}\includegraphics[width=0.2\textwidth]{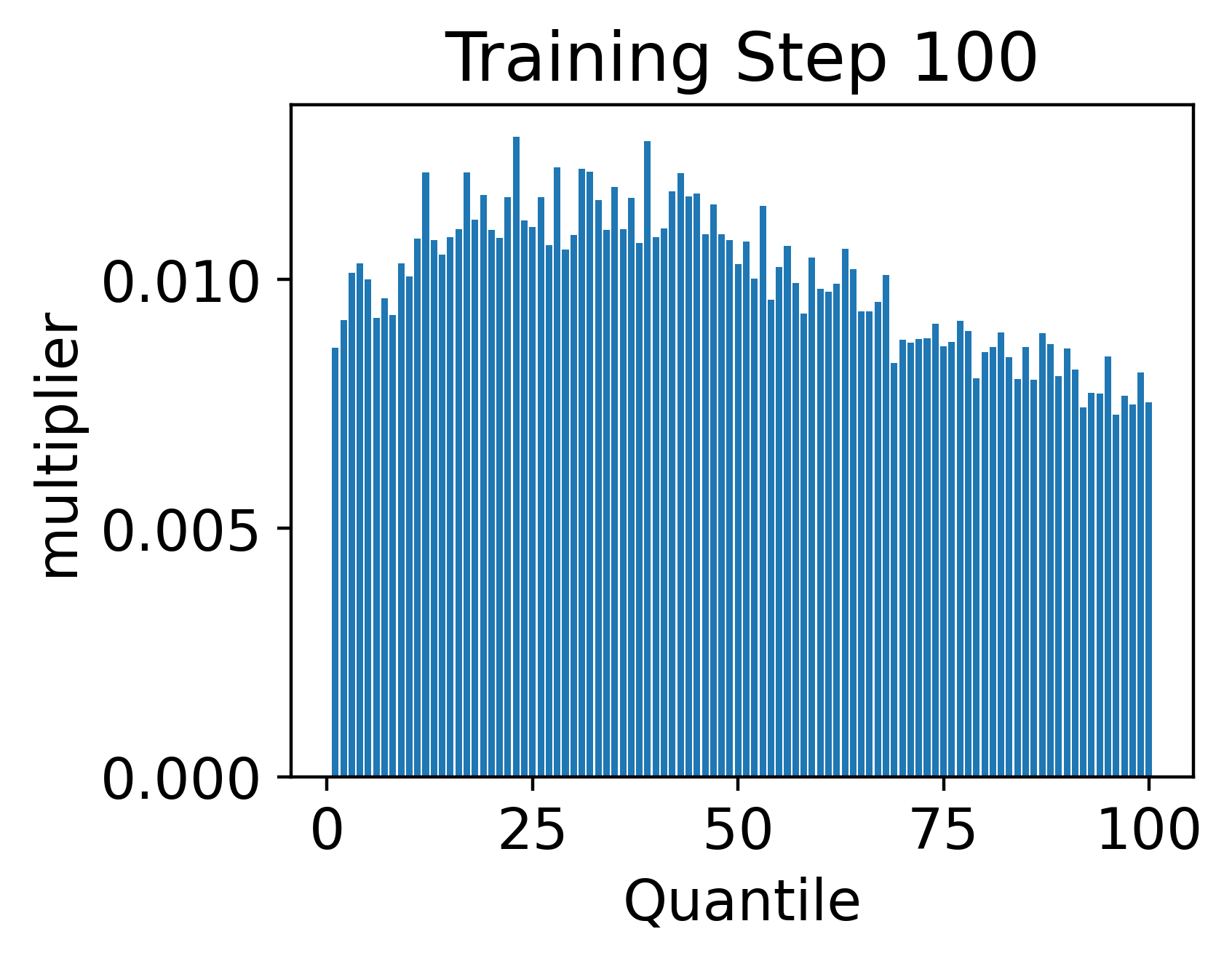}}
    \subfloat[Learned t=1000]{\label{fig:trimmed2}\includegraphics[width=0.2\textwidth]{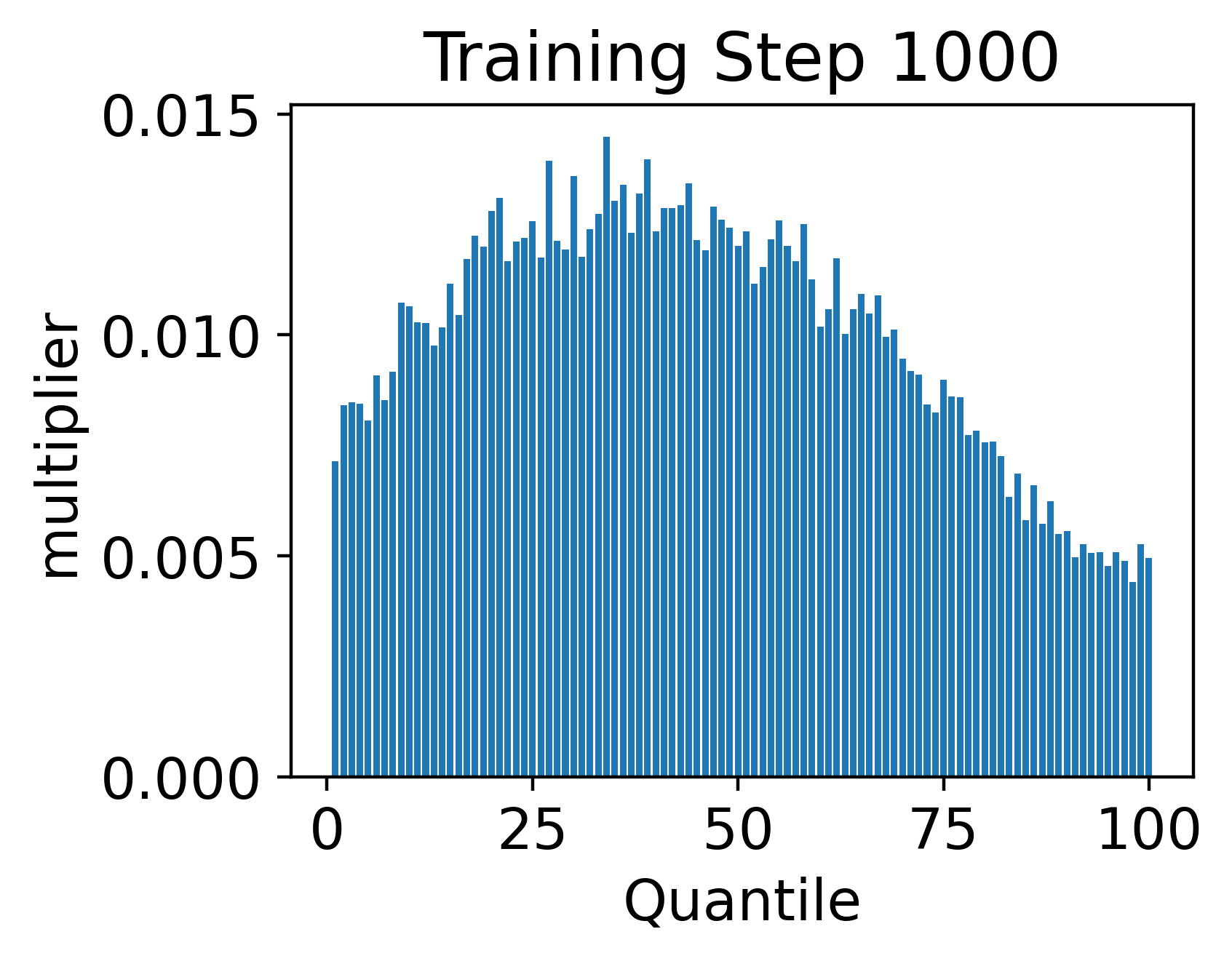}}
    \subfloat[Learned t=10000]{\label{fig:trimmed5}\includegraphics[width=0.2\textwidth]{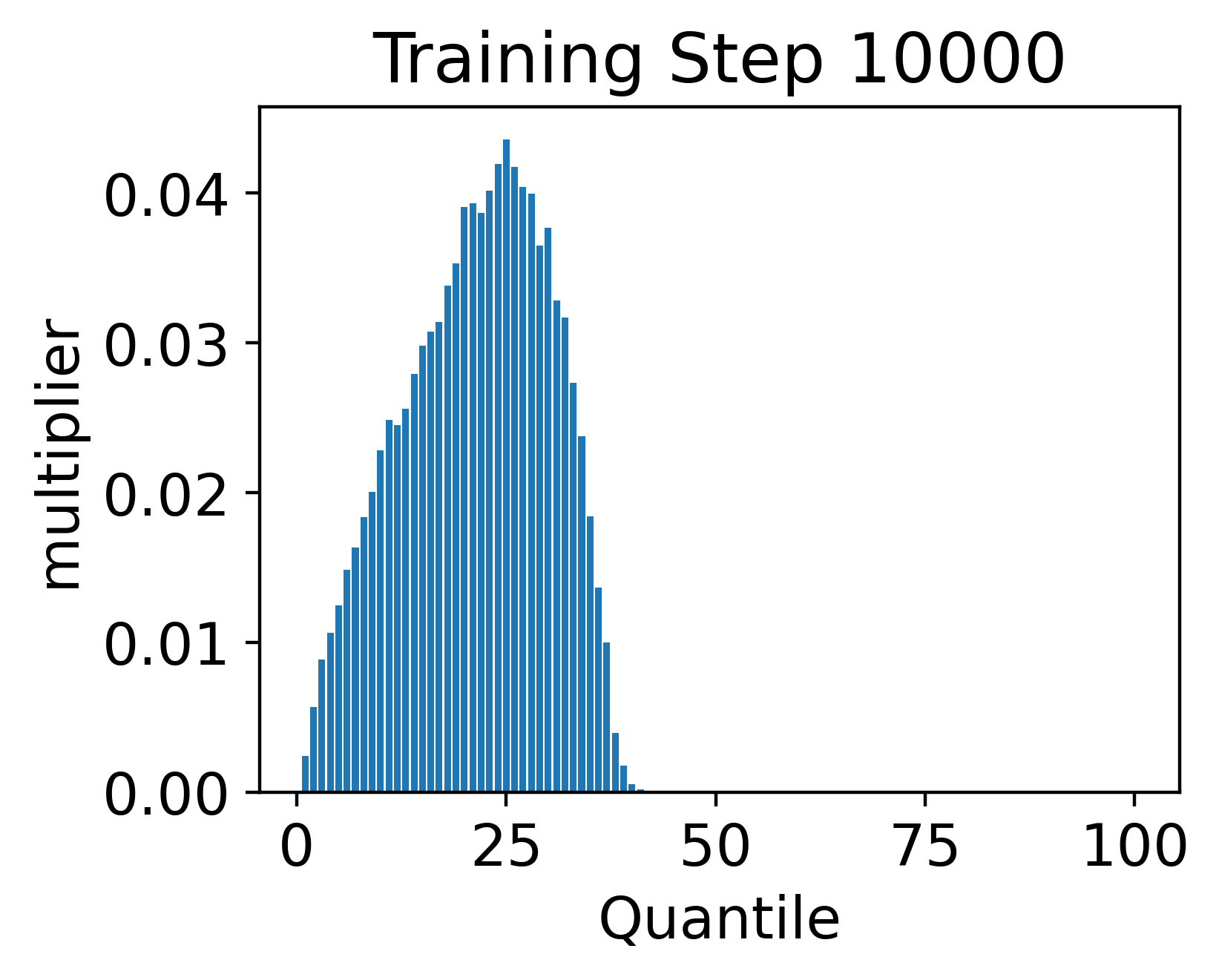}}
    \subfloat[Learned t=20000]{\label{fig:trimmed6}\includegraphics[width=0.2\textwidth]{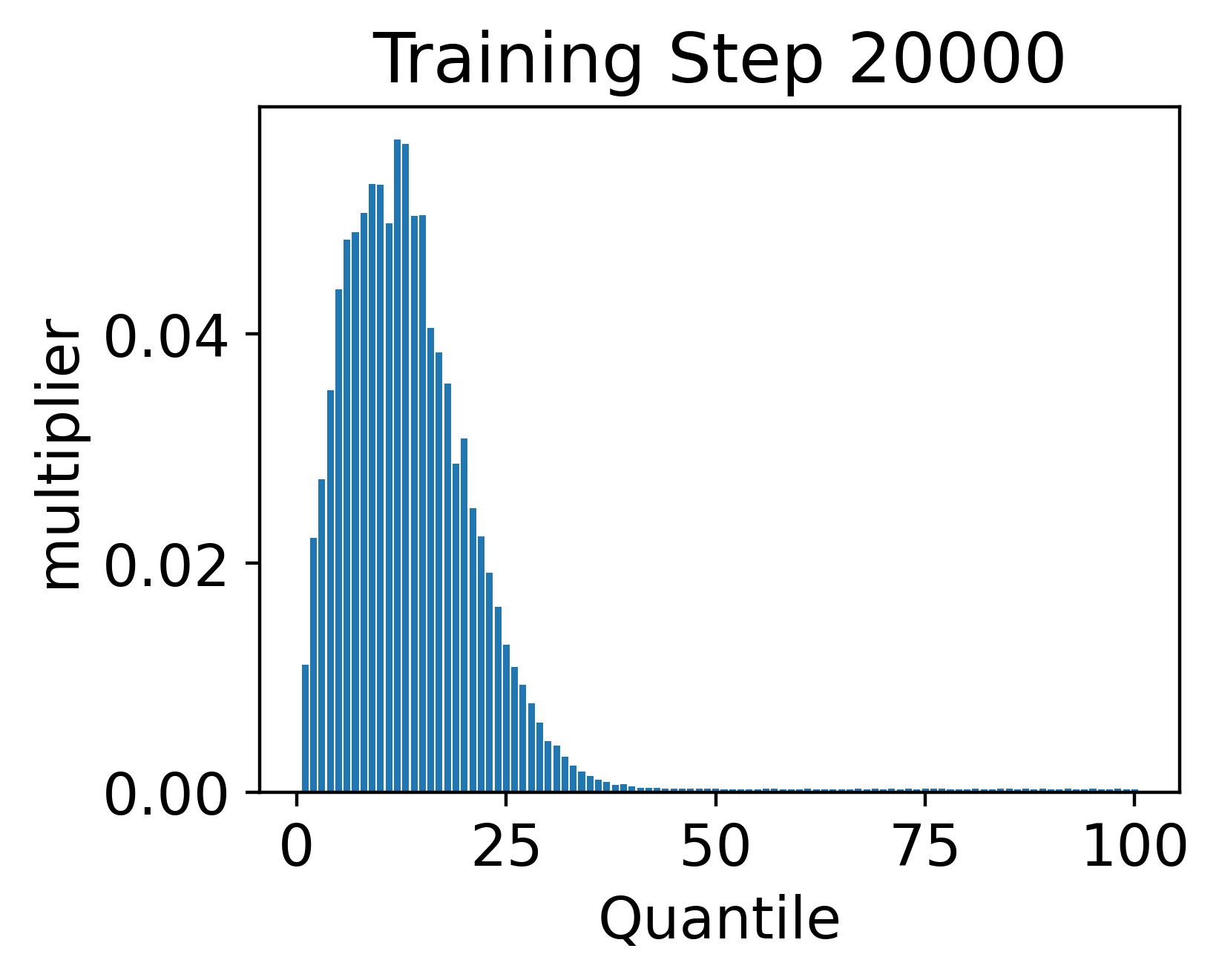}}
    \caption{Visualizations of the mini-batch risk functional learned when optimizing for the trimmed risk (at $\alpha = 0.1$) of the CIFAR-10 dataset, using a Resnet-18 model. Subfigure \ref{fig:trimmed} indicates the dataset-level risk functional we are optimizing for, and the remaining figures are the learned risk functional at different training time steps (indicated by the subfigure caption). The y-axis indicates the multiplier applied to the corresponding quantile (sorted by magnitude) of each mini-batch. The left side of each plot corresponds to the largest loss in the mini-batch, and the right side of each plot corresponds to the smallest loss in the mini-batch. The learned risk functional learns a warm-up schedule, and although initially focusing most on the middle quantiles of losses (as expected because of the trimmed risk functional), the learned mini-batch risk functional focuses more on the higher loss samples later in training.}
    \label{fig:Trimmed_RiskFns}
\end{figure*}

Interestingly, in all cases,
the learned mini-batch risk functional is either 
the best performing, or competitive with the best performing method, 
both in terms of loss and accuracy.
Table~\ref{tab:learned_riskfns_acc} highlights the importance
of warm-starting models when fine-tuning to specific 
risk functionals. 
And because our learned risk functionals are convex combinations
of mini-batch losses, sorted from highest to lowest loss,
there is a natural interpretation. 
Figures~\ref{fig:CVAR_RiskFns}, 
\ref{fig:ICVAR_RiskFns}, and \ref{fig:Trimmed_RiskFns}, 
show the learned risk functionals from 
multiple points throughout training.
We see that the learned risk functionals 
develop a curriculum that 

where we see that the learned risk functionals first 
optimizes the model in a sort of warm-starting phase 
where most samples are treated equally. 

However, as training further progresses, 
the learned risk functionals continue to become 
more and more specific to the setting. 
This automatic discovery of a warm-up period is visible in all 
of our learned risk functionals.
Another notable finding is that the final form of the 
learned mini-batch risk functionals are surprisingly different
from the underlying risk functional that they are optimizing for.

Here we further analyze the learned risk functionals 
obtained from optimizing for 
CVaR, ICVaR, and Trimmed Risk. 
In Figures~\ref{fig:CVAR_RiskFns}, \ref{fig:ICVAR_RiskFns}, 
and \ref{fig:Trimmed_RiskFns}, 
the leftmost plot shows the risk functional 
that we are optimizing for (at the \emph{dataset} level) 
in the same representation that
we are using to display our learned mini-batch risk functions. 
The remainder of each row shows the \emph{mini-batch} 
learned risk functional as it changes throughout training. 

Figure~\ref{fig:CVAR_RiskFns} details the risk functionals
resulting from optimizing for the CVaR risk functional. 
The CVaR risk functional focuses only on the top $\alpha$ 
(in this case $\alpha = 0.1$) percentile of highest losses. 
Or in other words,
we want a model that has minimum loss on the worst (or ``hardest'')
$\alpha$ percentile of the dataset. 
If we train a model
using CVaR as the risk functional at the batch level,
then we succeed in lowering the CVaR loss, 
but suffer a significant loss in overall accuracy
($\sim 68$\% accuracy down from $\sim 91$\%). 
Introducing warm starting to this procedure further improves 
both the loss (by about $7.5\%$) 
and the accuracy (by about $16\%$ relative) of the learned model. 
However, if we allow the mini-batch risk function to be learned, 
we get further improvement in both the loss and the accuracy
(relative improvements over the risk function of $10\%$ and $25\%$ respectively). 
Inspecting the learned mini-batch risk function (Figure~\ref{fig:CVAR_RiskFns}), 
we see that near the end of the training, 
the gradients from only the highest loss samples are kept, 
although not treated equally 
(with the most weight given to the highest loss sample, 
tapering off from there). 
Again, 
we see that the model has learned a sort of warm-up schedule, 
where earlier in training, all samples are treated relatively similarly, 
and then the weighting is slowly shifted 
towards the high-loss samples as training progresses. 
Of further note is the smoothness from which the learned mini-batch risk function 
transitions out of the warm-up phase, 
vs a more sudden transition caused by hand-engineered warm-up schedules. 

We see similar results when learning mini-batch risk functions for the ICVaR risk.
In this case, 
we are again using $\alpha = 0.1$, 
meaning that the loss we want to minimize is the loss on the 
lowest loss (``easiest'') 10\% of data. 
Namely, 
we again see the best-performing method being the learned risk functional
both in terms of loss and accuracy 
(with relative improvements over the risk functional with and without warm-up of 
$2\%$ and $99\%$ respectively). 
But more interestingly, 
we find that the learned risk functional responsible for this reduction 
in loss, 
while also improving accuracy, 
differs drastically from the ICVaR risk functional,
as shown in Figure~\ref{fig:ICVAR_RiskFns}. 
Again, warm-up schedule behavior is shown, 
and initially, more weight is placed on the lowest-loss samples
as we would expect. 
However, as training progresses, 
more weight is placed on the middle quantiles of losses. 
This seems to be in light of the fact that 
using just the lowest $10\%$ of losses in each mini-batch 
to train with yields an extremely poor model 
(${\sim}17\%$ accuracy as shown in Table~\ref{tab:learned_riskfns_acc}). 
Therefore, information from higher-loss samples is 
still beneficial to model optimization and risk reduction, 
even when only focusing on the ``easiest'' subset of the data. 

The learned mini-batch risk functional for optimizing the trimmed risk
again starts with a warm-up-like period before converging on a 
trimmed-risk-like risk functional, but, also again, biased towards the higher losses. 
We continue to use $\alpha = 0.1$, indicating that we discard the loss 
from the highest and lowest $10\%$ of the data. 
In all of the learned risk functionals, 
the transition point between samples that are given high vs low weight 
is much smoother than that of the risk functional being optimized. 

\subsection{Learning Mini-Batch Risk Functionals for Label Noise}
\label{sec:learningWLabelNoise}

\begin{table}[htb]
\centering
\caption{This table compares learning among label noise when also given access to a noise-free validation set of 1,000 samples (2\% of the training set). Column 1 indicates the percentage of the training labels that were randomly assigned, column 2 indicates the baseline test accuracy of training with the expected value, and column 3 shows the accuracy when training with a learned risk functional. We also trained a model only on the clean data, which achieves an accuracy of 61.21\% (1.712).}
\label{tab:labelNoise_withCleanVal}
\begin{tabular}{@{}ccc@{}}
\toprule
\% Random Labels & \begin{tabular}[c]{@{}c@{}}Expected\\ Value\end{tabular} & \begin{tabular}[c]{@{}c@{}}Learned\\ Risk Function\end{tabular} \\ \midrule
0\%   & \textbf{91.25 (0.241)}                                 & \textbf{91.08 (0.277)}                                          \\
5\%   & 80.98 (0.161)                                          & \textbf{82.65 (0.549)}                                           \\
20\%  & 71.75 (0.909)                                           & \textbf{79.02 (0.4807)}                                          \\
50\%  & 60.18 (2.94)                                             & \textbf{68.75 (1.42)}                                          \\ \bottomrule
\end{tabular}
\end{table}

\begin{table}[htb]
\centering
\caption{This table compares learning among label noise, but without access to a noise-free validation set. Column 1 indicates the percentage of the training labels that were randomly assigned, column 2 indicates the baseline test accuracy of training with the expected value, column 3 shows the accuracy when training with a learned mini-batch risk functional, and column 4 shows the oracle accuracy. The oracle is given information such as the optimal risk functional and corresponding parameters to use, which is typically not available.}
\label{tab:labelNoise_NoCleanVal}
\begin{tabular}{@{}cccc@{}}
\toprule
\begin{tabular}[c]{@{}c@{}}Random\\ Labels\end{tabular} & \begin{tabular}[c]{@{}c@{}}Expected\\ Value\end{tabular} & \begin{tabular}[c]{@{}c@{}}Learned \\ Risk Function\end{tabular} & Oracle \\ \midrule
0\%   & \textbf{91.25 (0.241)} & \textbf{91.08 (0.277)} & - \\
5\%   & 80.62 (0.498) & \textbf{81.71 (0.289)} & 84.07 (0.189) \\
20\%  & 68.46 (0.394) & \textbf{77.85 (0.557)} & 82.31 (0.376) \\
50\%  & 60.72 (2.46) & \textbf{69.54 (1.23)} & 76.86 (0.311) \\ \bottomrule
\end{tabular}
\end{table}

\begin{figure*}[htb]
    \centering
    \subfloat[Learned t=100]{\label{fig:ln50}\includegraphics[width=0.2\textwidth]{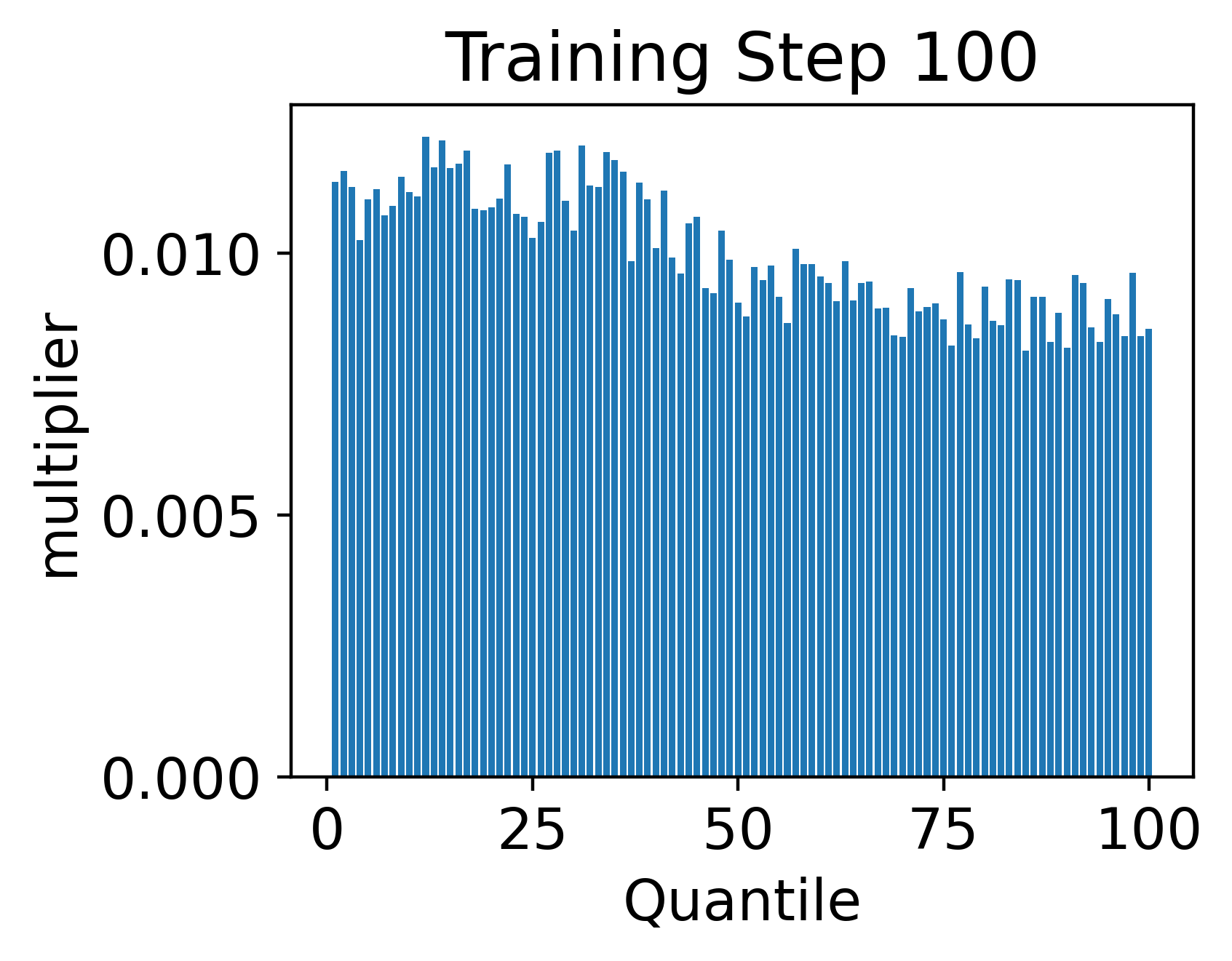}}
    \subfloat[Learned t=1000]{\label{fig:ln500}\includegraphics[width=0.2\textwidth]{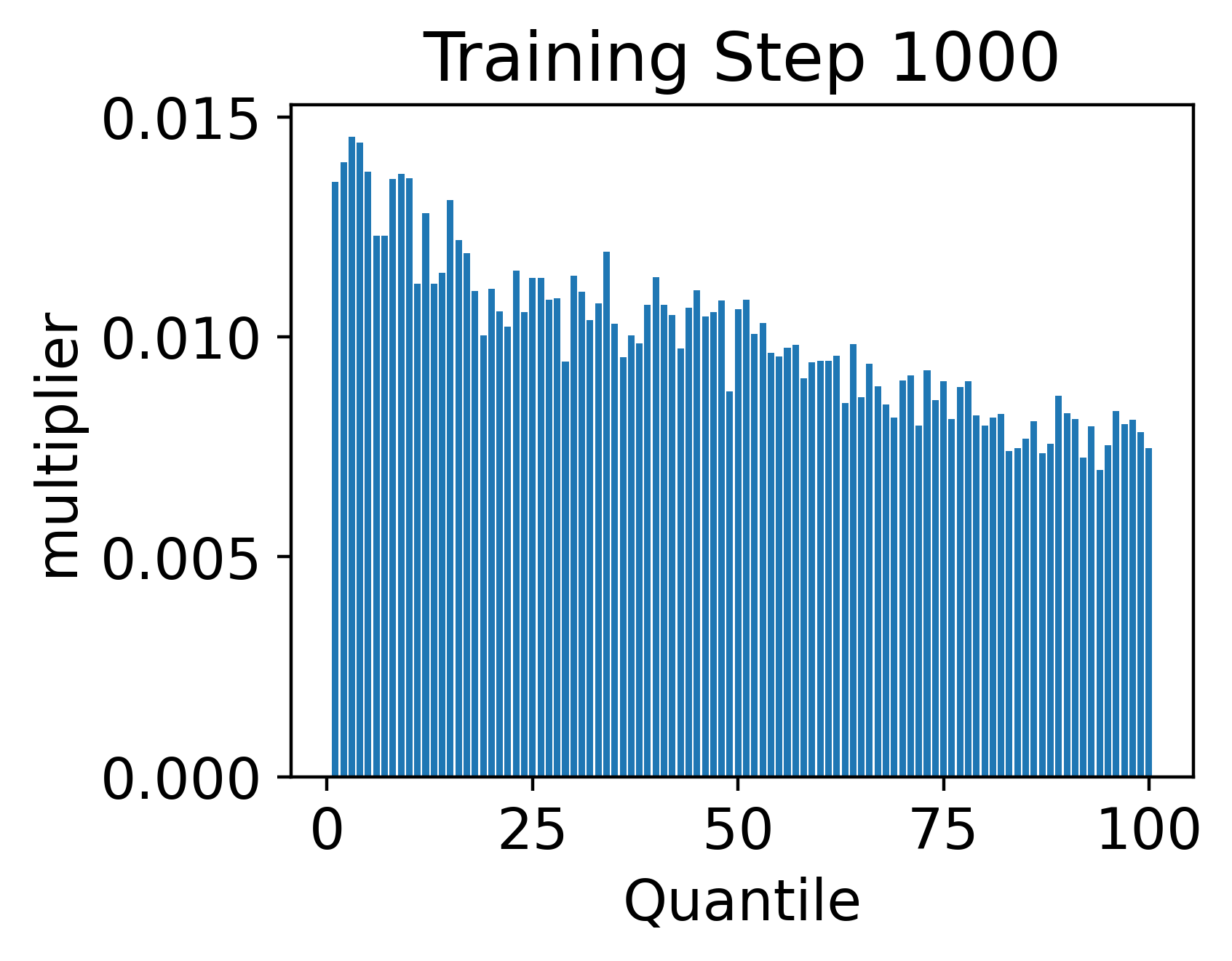}}
    \subfloat[Learned t=2000]{\label{fig:ln501}\includegraphics[width=0.2\textwidth]{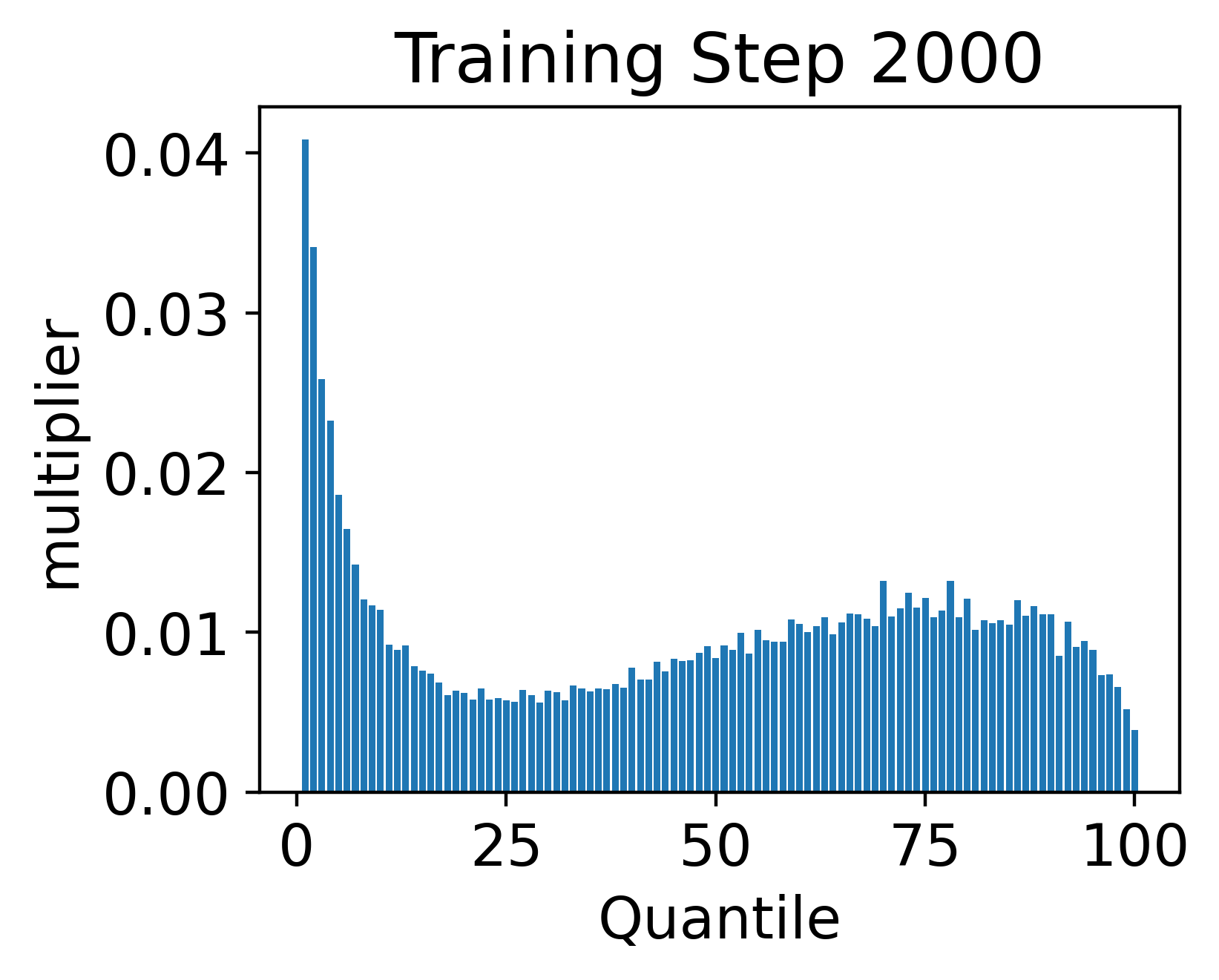}}
    \subfloat[Learned t=10000]{\label{fig:ln502}\includegraphics[width=0.2\textwidth]{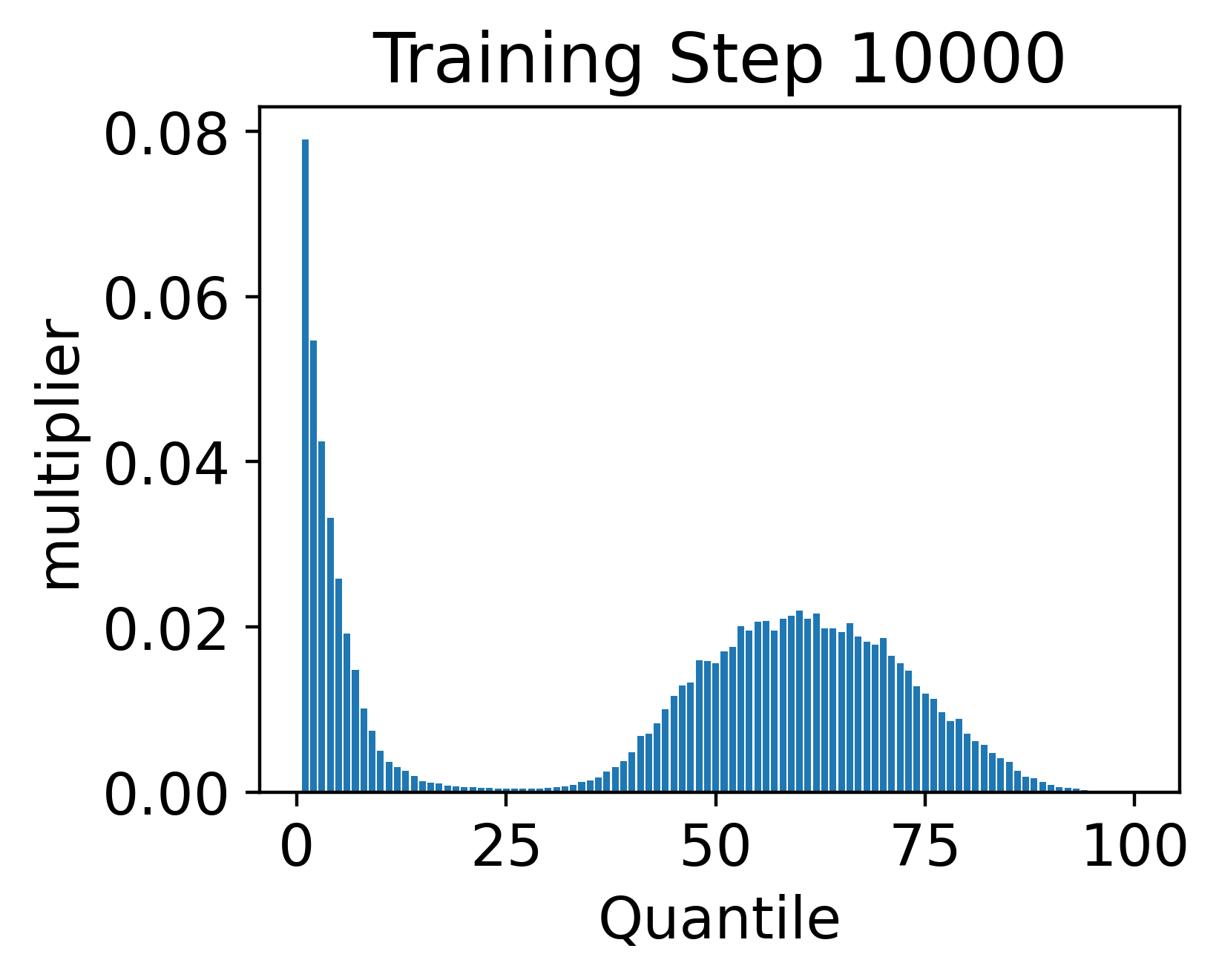}}
    \subfloat[Learned t=20000]{\label{fig:ln503}\includegraphics[width=0.2\textwidth]{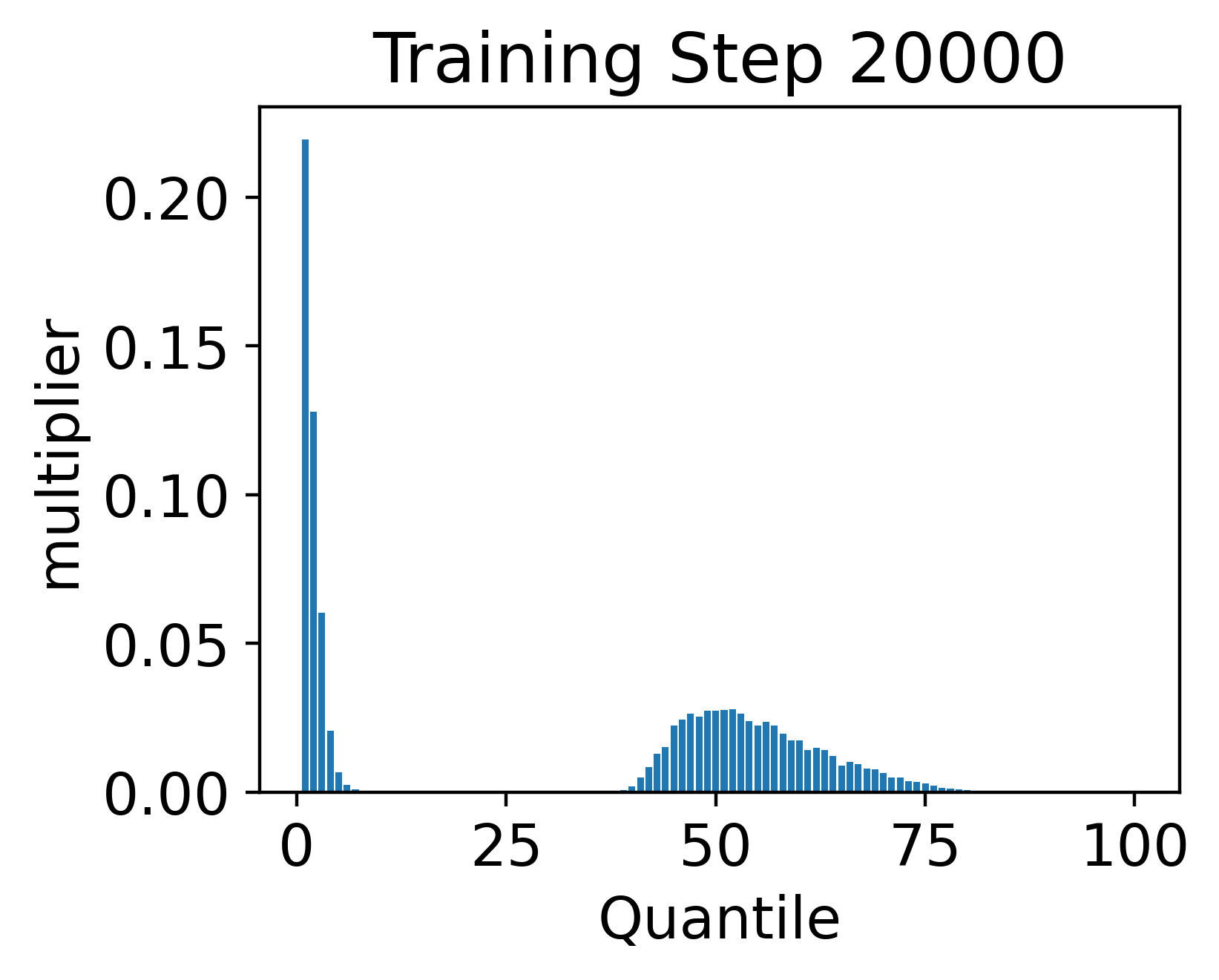}}
    \caption{Visualizations of the mini-batch risk functional learned when optimizing for the expected value of the validation loss of the CIFAR-10 dataset with 50\% random labels, using a Resnet-18 model. Each figure represents the learned mini-batch risk functional at different training time steps (indicated by the subfigure caption). The y-axis indicates the multiplier applied to the corresponding quantile (sorted by magnitude) of each batch. The left side of each plot corresponds to the largest loss in the batch, and the right side of each plot corresponds to the smallest loss in the batch. The learned risk functional learns a warm-up schedule, and then begins to down weight most of the high-loss samples (except for the very highest losses) and focus more on the lower-loss quantiles. Surprisingly, even late in training, the focus on the lower quantiles remains, but so does the emphasis on a small number of high-loss samples.}
    \label{fig:labelNoiseRiskFns}
\end{figure*}

Now that we have demonstrated the effectiveness of our method
in finding useful mini-batch risk functionals, 
we explore its applicability to a specific distribution shift: label noise. 
In this section, 
we introduce different levels of label noise to the CIFAR10 dataset, 
and then explore the different models learned with and without 
a mini-batch learned risk functional. 
To induce noise, 
we randomly select a percentage of the training data, 
and then reassign labels uniformly at random. 
We first explore the models learned 
when given access to the noisy data along with a small, 
noise-free, subset of $1,000$ samples ($2\%$ of the training set). 
Then, more interestingly, 
we demonstrate that the performance of our method 
suffers only very minor degradation when a noise-free subset is not present. 

Table~\ref{tab:labelNoise_withCleanVal} shows the accuracy 
when training a model with and without a learned mini-batch risk functional, 
given a noise-free validation set of 1,000 samples. 
As the noise becomes more drastic, 
the benefit of a learned risk functional increases. 
Notably, 
even when no label noise is present, 
using a learned mini-batch risk functional reaches \emph{equal} accuracy 
to that of a model trained normally. 
At 5\% label noise, a learned mini-batch risk function improves accuracy by ${\sim}1.7\%$ absolute, 
whereas at 50\% label noise, this improvement is ${\sim}8.6\%$ absolute. 

Conventional wisdom is that the correct thing to do 
when learning with a known level of 50\% label noise
is to train a model with 
the ICVaR functional with $\alpha=0.5$. 
This would tell the model to focus only on the lowest 50\% of losses in the dataset, 
therefore ignoring the mislabeled 50\% 
(with the assumption that the mislabeled data points have 
higher loss than correctly labeled data points). 
However, as indicated from the results in Table~\ref{tab:learned_riskfns_acc}, 
warm-starting is critical to training effective models with ICVaR
(and warm-starting makes the aforementioned assumption likely to hold). 
As the results from Table~\ref{tab:learned_riskfns_acc} further demonstrate, 
simply training with this risk functional applied at the mini-batch does not 
always produce the best results. 
Although our method \emph{does not} have access to information such as the 
amount of noise in the dataset,
or even the type of distribution shift present, 
we compare against an \emph{Oracle} model that is given all of this information
and therefore can be optimally trained. 
These results are added under 
the Oracle column in Table~\ref{tab:labelNoise_NoCleanVal}.

The mini-batch risk functional learned when 50\% label noise is present is
shown in Figure~\ref{fig:labelNoiseRiskFns} 
(note that at 50\% random labels, 
there is a chance that each label is 
randomly assigned the proper label - 
resulting in a true level of label noise around 45\%). 
Interestingly, we again see the automatically learned warm-up behavior, 
but the learned mini-batch risk functional resembles the ICVaR functional only in part. 
While the latter percentiles of loss are similar (roughly the 50th to 75th percentile), 
it places high emphasis on a few of the highest-loss samples. 
This emphasis on high-loss samples is surprising, 
even though the emphasis on them becomes less throughout training.

\section{Related Work}
\label{relatedwork}

Several studies have focused on understanding how to optimize for 
risk functionals other than the 
average~\citep{duchi2021learning, duchi2022distributionally}. 
Furthermore, \citet{leqi2019human} showed that the minimization 
of risk functionals \emph{other than} the expected loss is beneficial 
with respect to some desiderata such as fairness, 
and for better aligning with a more human-like risk measure.
In \citet{li2020tilted}, 
a hand-engineered functional for determining the weight for each sample
in a mini-batch is presented, 
and display its effectiveness in enforcing fairness, mitigating outliers, and handling class imbalance. 
They show that this sample weighting scheme can adapt to new applications such as simultaneously addressing outliers and promoting fairness.

While we present a method for learning risk functionals in any setting, 
one related application that has received much attention is 
learning with noisy labels. 
Although we focus on the more general setting, here we include some of the most related work in this subfield. 
After it was proven that a deep neural network 
with a modified loss function 
for noisy data can approach the 
Bayes optimal classifier~\citep{manwani2013noise, ghosh2017robust},
many efforts at hand-engineering more robust loss functionals followed. 
These include generalized gross entropy~\citep{zhang2018generalized}, 
Bi-tempered loss~\citep{amid2019robust}, 
and 0-1 loss surrogate loss functionals that can ignore samples,
based on a threshold \citet{lyu2019curriculum}.
While these have shown benefits in specific settings, they have not been adopted for more general use.
Our work extends this by learning a sample weighting function instead
of relying on hand engineering. 

Estimation of a noise transition matrix is also common in 
learning with noisy labels. 
The noise transition matrix is then 
used to transform model outputs from a noiseless to noisy label 
domain~\citep{patrini2017making, hendrycks2018using, yao2020dual, yang2021estimating}. 
Similarly, \citet{xiao2015learning, han2018masking, yao2018deep} 
build network architectures specific for learning among label noise. 
These methods focus on mapping from model output to noise-adapted posterior, 
whereas we learn a reduction function 
that maps from batch losses to reduced batch loss.

Another interpretation of our method is that it is a form of importance weighting.
\citet{liu2015classification} apply importance weighting 
to learning with noisy labels. 
They present a method of estimating the target and 
source distribution labels using kernel density estimation. 
\citet{wang2017multiclass} also apply importance weighting in this setting, 
by modifying the updates according to an estimate 
of the proportion of randomly labeled examples. 
\citet{chang2017active} follow this setting as well,
but their updates are modified according to the 
variance in predicted probabilities of the correct class 
across mini-batches. 
Again, our method goes beyond these by learning a weighting function
from a much larger function class (i.e. more expressive function classes), 
and in that it operates as a loss reduction strategy leveraging only information 
available in a batch of data. 

\citet{Jenni_2018_ECCV} introduce bi-level optimization that learns 
a function that assigns a scalar weight to the batch loss. 
More specifically, given a clean validation set, 
each mini-batch is weighted according to how well its gradient aligns with 
one calculated on clean data. 
In contrast, our method learns to weight each sample in a batch before reduction
and does not require clean data. 
Meta-weight-net~\cite{shu2019meta} learns a function that 
scales the loss of every sample in a batch 
(independent of the other samples in the batch - 
a sample producing the same loss in multiple batches 
will be scaled equally to the same value).
\citet{ren2018learning} also learn a reduction function that maps 
from batch losses to reduced batch loss
via a meta-learning-inspired update.
Our method differs from these two works by
being sensitive to the relative performance of samples within each batch 
(achieved by enforcing sorting and then scaling the loss of each sample
with respect to their ranking in the batch - 
the same loss in different batches are treated differently) 
instead of the loss value itself, 
and by allowing flexibility in the number of inner steps 
(i.e. more exact approximations). 

\citet{pmlr-v97-shen19e} minimize the trimmed loss by 
alternating between selecting true-labeled examples
and retraining a model. 
At each round, 
they keep only a fraction of small-loss examples to retrain the model 
during the next round. 
In contrast, 
we learn our risk functionals in a single shot. 
Lastly, as we will demonstrate in Section~\ref{sec:learningWLabelNoise}, 
our method does not depend on having a clean validation set. 
We are not aware of any work that presents a general-use method 
for learning mini-batch risk functionals. 
\section{Conclusion}

We introduced a meta-learning-based method of learning 
an interpretable mini-batch risk functional
during model training, in a single shot,
that both outperforms hand-engineered risk functionals 
and discovers useful mini-batch risk functionals in settings 
where they are unknown a priori. 
When optimizing for various risk functionals, 
the learned mini-batch risk functions lead to risk reduction of up to 10\% 
over hand-engineered ones. 
Then in the presence of label noise, 
where risk functionals can be useful but the right one is unknown, 
our method improved over baseline by 
over 14\% relative ($\sim$9\% absolute) among 50\% random labels, 
even when given \emph{no} noise-free data.
Our learned mini-batch risk functions 
are restricted to taking a convex
combination of mini-batch loss quantiles, 
and are therefore clearly interpretable. 
We analyzed the learned mini-batch risk functionals at different points through training, 
and found that they learned curriculums (including warm-up periods), 
and that their final form can be surprisingly different from the 
underlying risk functional that they optimized for.

\clearpage

\bibliography{references}
\bibliographystyle{icml2023}

\end{document}